\documentclass[11pt]{article}

\usepackage[final]{acl}

\usepackage{times}
\usepackage{latexsym}

\usepackage[T1]{fontenc}
\usepackage[utf8]{inputenc}

\usepackage{microtype}

\usepackage{inconsolata}

\usepackage{graphicx}
\usepackage{amsmath}
\usepackage{url}

\usepackage{booktabs}   % professional table rules
\usepackage{xcolor}     % colour support
\usepackage{xspace}     % smart spacing after macros

\usepackage{times}
\usepackage{latexsym}
\usepackage{enumitem}
\usepackage{listings}
\usepackage{url}
\usepackage[T1]{fontenc}
\usepackage[utf8]{inputenc}

\usepackage{microtype}

\usepackage{inconsolata}

\usepackage{longtable}
\usepackage{tabularx}
\usepackage{graphicx}

\usepackage{caption}
\usepackage{subcaption}

\usepackage{colortbl}
\usepackage{makecell}
\usepackage{multirow}
\usepackage{supertabular}
\usepackage{placeins}

\usepackage{subfiles}
\usepackage{enumitem}

\usepackage{algorithm}
\usepackage{algpseudocode}
\usepackage{amsmath}

\usepackage{cleveref}

\setlist[itemize]{leftmargin=1em}  % smaller left margin = more text on one line

\usepackage{booktabs}

\usepackage{tcolorbox}
\newcounter{promptno}[section]
\newlength\mystoreparindent

\usepackage{amssymb}

\DeclareUnicodeCharacter{25A2}{$\square$}
\usepackage{color}

\newcolumntype{T}{>{\ttfamily}l}

\tcbuselibrary{skins}

\definecolor{a-colour}{RGB}{135,74,175}
\definecolor{b-colour}{RGB}{243,145,9}
\definecolor{gm-colour}{RGB}{128,128,128}
\definecolor{wordlegreen}{rgb}{0.4,0.79,0.17}
\definecolor{wordlered}{rgb}{0.86,0.31,0.29}
\definecolor{wordleyellow}{rgb}{1,1,0}

\def \dcolwidth {3cm}
\def \skipcolwidth {0.8cm}
\def \arccorner {3mm}
\def \disttocorner {6pt}

\newtcolorbox[use counter=nbubbles]{a-gm}[1]{
    colback=a-colour!10!white,
    colframe=a-colour,
    fonttitle=\bfseries\tiny,
    fontupper=\footnotesize,
    title={#1},
    sharp corners=west,
    arc=\arccorner,
    width=\dcolwidth,
    left skip=0cm,
    top=0pt,
    bottom=0pt,
    left=0pt,
    right=\disttocorner,
    before={\vspace{-0.1cm}},
    boxrule=0.5pt,
    enhanced,
    attach boxed title to top left={yshift=-0.1mm},
    boxed title style={size=small,colback=a-colour},
    overlay unbroken and first = {
    \node[text width=0.4cm,draw=black,line width=0.1mm,align=center,gray] at (-0.5,0.2) {\footnotesize \thetcbcounter};
  }
}

\newtcolorbox[use counter=nbubbles]{gm-a}[1]{
    colback=gm-colour!5!white,
    colframe=gm-colour,
    fonttitle=\bfseries\tiny,
    fontupper=\footnotesize,
    title={#1},
    sharp corners=east,
    arc=\arccorner,
    width=\dcolwidth + \skipcolwidth,
    left skip=\skipcolwidth,
    top=0pt,
    bottom=0pt,
    left=\disttocorner,
    right=0pt,
    before={\vspace{-0.1cm}},
    boxrule=0.5pt,
    halign title=flush right,
    enhanced,
    attach boxed title to top right={yshift=-0.1mm},
    boxed title style={size=small,colback=gm-colour},
    overlay unbroken and first = {
    \node[anchor=north east,text width=0.4cm,draw=black,line width=0.1mm,align=center,gray] at (-0.9,0.55) {\footnotesize\thetcbcounter};}
}

\newtcolorbox[use counter=nbubbles]{b-gm}[1]{
    colback=b-colour!10!white,
    colframe=b-colour,
    fonttitle=\bfseries\tiny,
    fontupper=\footnotesize,
    title={#1},
    sharp corners=east,
    arc=\arccorner,
    width=\dcolwidth + \skipcolwidth + \dcolwidth +  \skipcolwidth,
    left skip=\dcolwidth + \skipcolwidth + \skipcolwidth,
    top=0pt,
    bottom=0pt,
    left=\disttocorner,
    right=0pt,
    before={\vspace{-0.1cm}},
    boxrule=0.5pt,
    halign title=flush right,
    enhanced,
    attach boxed title to top right={yshift=-0.1mm},
    boxed title style={size=small,colback=b-colour},
    overlay unbroken and first = {
    \node[anchor=north east,text width=0.4cm,draw=black,line width=0.1mm,align=center,gray] at (-4.7,0.55) {\footnotesize\thetcbcounter};}
}

\newtcolorbox[use counter=nbubbles]{gm-b}[1]{
    colback=gm-colour!5!white,
    colframe=gm-colour,
    fonttitle=\bfseries\tiny,
    fontupper=\footnotesize,
    title={#1},
    sharp corners=west,
    arc=\arccorner,
    width=\dcolwidth + \skipcolwidth + \dcolwidth,
    left skip=\dcolwidth + \skipcolwidth,
    top=0pt,
    bottom=0pt,
    left=0pt,
    right=\disttocorner,
    before={\vspace{-0.1cm}},
    boxrule=0.5pt,
    enhanced,
    attach boxed title to top left={yshift=-0.1mm},
    boxed title style={size=small,colback=gm-colour},
    overlay unbroken and first = {
    \node[anchor=north east,text width=0.4cm,draw=black,line width=0.1mm,align=center,gray] at (-3.9,0.55) {\footnotesize\thetcbcounter};}
}

\newtcolorbox[use counter=nbubbles]{gm-gm}[1]{
    colback=gm-colour!5!white,
    colframe=gm-colour,
    fonttitle=\bfseries\tiny,
    fontupper=\footnotesize,
    title={#1},
    sharp corners,
    width=\dcolwidth + \dcolwidth + \skipcolwidth + \skipcolwidth,
    leftright skip=\dcolwidth,
    top=0pt,
    bottom=0pt,
    left=0pt,
    right=0pt,
    before={\vspace{-0.1cm}},
    boxrule=0.5pt,
    halign title=center,
    enhanced,
    attach boxed title to top center={yshift=-0.1mm},
    boxed title style={size=small,colback=gm-colour},
    overlay unbroken and first = {
    \node[anchor=north east,text width=0.4cm,draw=black,line width=0.1mm,align=center,gray] at (-3.1,0.55) {\footnotesize\thetcbcounter};}
}

\renewcommand{\topfraction}{0.95}
\renewcommand{\bottomfraction}{0.7}
\renewcommand{\textfraction}{0.05}
\renewcommand{\floatpagefraction}{0.8}
\renewcommand{\dbltopfraction}{0.95}
\renewcommand{\dblfloatpagefraction}{0.8}

\title{Language Equality has a Price: A Systematic Investigation \\ of Multi-turn LLM Performance for EU-24+}

\author{%
Sherzod Hakimov, Karl Osswald, Jelle Psurek, Eszter Bukovszky\\
\textbf{A. Altar Lüser, David Schlangen${^\mathbf{1}}$ }
\thanks{$\;$ Contributions: SH initiated and managed the project, wrote the publication, coded the game localization pipeline and ran all experiments. KO ran the open-weight model experiments, verified Chinese and Greek translations and applied fixes that improved all games. JP set up the manual translation and verification pipeline and verified the German language. EB set up the initial pipeline for localization and verified Hungarian language. AL verified the Turkish language. DS supervised the project, and edited the main part of the paper.}\\
Computational Linguistics, Department of Linguistics\\
University of Potsdam, Germany\\
$^{\mathbf{1}}$German Research Center for Artificial Intelligence (DFKI), Berlin, Germany\\
{\texttt{\{firstname.lastname\}@uni-potsdam.de}}
}

\begin{document}
\maketitle

\begin{abstract}
We evaluate large language models (LLMs) as language agents playing goal-directed dialogue games in self-play across 30 languages: the 24 official EU languages plus six others. Unlike static or preference-based evaluation, this paradigm is multi-turn, reference-free and programmatically scored, and because the game mechanics are language-agnostic it extends to a new language by localising a fixed set of prompt and word-list files. Evaluating nine open-weight and commercial LLMs, we find that no open-weight model covers the EU-24 well: in every official language both commercial systems outscore every open-weight model, and the two weakest average below 40 points across the EU-24. The commercial systems stay ahead even in languages with four orders of magnitude less public web text, showing that linguistic parity is achievable, but not from public crawls alone. A model's home region lifts it without closing the gap: Chinese is the strongest of all 30 languages for two Chinese-developed models, yet the best Chinese score of any model belongs to a US commercial system. Coverage is also not parity of service. Pooled over models and languages, the median non-English language costs 31\% more to run than English, and scores 10\% lower.
\end{abstract}

\section{Introduction}
\label{sec:intro}

\textit{``Every person may write to the institutions of the Union in one of the languages of the Treaties and must have an answer in the same language.''}
\hfill{\small(EU Charter of Fundamental Rights, Art.~41(4)~\cite{eucharter2012})}

\medskip

\noindent This commitment to equal standing becomes hollow when the AI systems that mediate such interactions perform unevenly across languages. Large language models are known to do exactly that \citep{DBLP:conf/emnlp/LaiNVMDBN23,DBLP:conf/acl/UstunAYKDOBSOKV24}, which affects who benefits from them. Tests of LLM ability must therefore cover every target language, and for the EU that means all languages of the Treaties. What a benchmark measures matters as much as which languages it covers. Real language use is dialogic: it consists of sustained, goal-directed interaction in which each contribution builds on the last, not of one-shot question answering. Assessing multilingual ability therefore calls for evaluation that is interactive and spans multiple capabilities in both high- and low-resource languages.

Existing multilingual evaluation falls short of this along four dimensions, on which Table~\ref{tab:rw-comparison} compares the main benchmarks. Most are \emph{single-turn}: the dominant approach translates static English test sets into many languages \citep{DBLP:conf/acl/SinghRFANVLMLSN25,DBLP:conf/emnlp/XuanYQZXFLXWGLJLLYDGLXJ25,DBLP:journals/corr/abs-2506-19468,DBLP:journals/corr/abs-2410-08928}, and native-source counterparts \citep{DBLP:conf/iclr/RomanouFSNSMACH25,DBLP:conf/nips/ZhangAGCB23,isbarov-etal-2025-tumlu} avoid translation artefacts but are single-turn as well. Many score a selection among given options rather than \emph{open-ended generation}. Few score adherence to formal constraints explicitly, and the multilingual \emph{instruction-following} benchmarks that do \citep{DBLP:journals/corr/abs-2410-15553,DBLP:journals/corr/abs-2503-07539} script their turns in advance rather than making them contingent on the model's own output. Finally, suites that do span \emph{multiple capabilities} probe each one in a separate subtask \citep{DBLP:conf/emnlp/HuangZHHLHY25,DBLP:conf/emnlp/AhujaDHORJNGSAB23}, so no single episode shows how those capabilities interact. Section~\ref{sec:related} details this landscape.

\providecommand{\cmark}{\checkmark}
\providecommand{\xmark}{$\times$}

\begin{table*}[t]
\centering
\small
\setlength{\tabcolsep}{4pt}
\resizebox{\textwidth}{!}{%
\begin{tabular}{l@{\hspace{6pt}}cccc}
\toprule
\textbf{Benchmark} & \textbf{Multi-turn} & \textbf{Open-ended generation} & \textbf{Instruction following} & \textbf{Multiple capabilities} \\
\midrule
Global MMLU \citep{DBLP:conf/acl/SinghRFANVLMLSN25} & \xmark & \xmark & \xmark & \xmark \\
MMLU-ProX \citep{DBLP:conf/emnlp/XuanYQZXFLXWGLJLLYDGLXJ25} & \xmark & \xmark & \xmark & \xmark \\
M3Exam \citep{DBLP:conf/nips/ZhangAGCB23} & \xmark & \xmark & \xmark & \xmark \\
INCLUDE \citep{DBLP:conf/iclr/RomanouFSNSMACH25} & \xmark & \xmark & \xmark & \xmark \\
MultiLoKo \citep{DBLP:journals/corr/abs-2504-10356} & \xmark & \cmark & \xmark & \xmark \\
MuBench \citep{DBLP:journals/corr/abs-2506-19468} & \xmark & \xmark & \xmark & \cmark \\
Eurolingua \citep{DBLP:journals/corr/abs-2410-08928} & \xmark & \xmark & \xmark & \cmark \\
PolyMath \citep{DBLP:journals/corr/abs-2504-18428} & \xmark & \cmark & \xmark & \xmark \\
OneRuler \citep{DBLP:journals/corr/abs-2503-01996} & \xmark & \cmark & \xmark & \xmark \\
BenchMAX \citep{DBLP:conf/emnlp/HuangZHHLHY25} & \xmark & \cmark & \cmark & \cmark \\
Multi-IF \citep{DBLP:journals/corr/abs-2410-15553} & \cmark & \cmark & \cmark & \xmark \\
XIFBench \citep{DBLP:journals/corr/abs-2503-07539} & \xmark & \cmark & \cmark & \xmark \\
IrokoBench \citep{DBLP:conf/naacl/AdelaniOAZAHOHBLCBSKMKY25} & \xmark & \cmark & \xmark & \cmark \\
TUMLU \citep{isbarov-etal-2025-tumlu} & \xmark & \xmark & \xmark & \xmark \\
Karde\c{s}-NLU \citep{senel-etal-2024-kardes} & \xmark & \xmark & \xmark & \cmark \\
ProverbEval \citep{DBLP:conf/naacl/AzimeTBCBAANYGtSSK25} & \xmark & \cmark & \xmark & \xmark \\
\midrule
\textbf{Ours} & \cmark & \cmark & \cmark & \cmark \\
\bottomrule
\end{tabular}%
}
\caption{Comparison of multilingual LLM benchmarks along the four dimensions identified as missing in the preceding discussion: evaluation spans multiple dialogue turns (\textbf{Multi-turn}); responses are produced as free-form text rather than selected from options (\textbf{Open-ended generation}); adherence to formal constraints is explicitly scored (\textbf{Instruction following}); multiple distinct capabilities are covered (\textbf{Multiple capabilities}). \cmark~= yes, \xmark~= no.}
\label{tab:rw-comparison}
\end{table*}

We address these gaps with a multilingual, interactive benchmark in which models are evaluated as agents playing goal-directed dialogue games in self-play \citep{chalamalasetti2023clembench,DBLP:journals/corr/abs-2507-08491}. The game mechanics are language-agnostic, so a new language requires only localised game files (prompts, parsing rules, word lists) and no reference answers. This turns full institutional coverage from a data-collection project into a localisation task, which we use to evaluate interactive language use across all 24 official EU languages. It also lets us ask what static benchmarks cannot: not just how well a model performs in a language, but at what cost. We pair performance with the tokens and dollars it costs per language, and relate both to available web text and the economic weight of the speaker community.

We make three contributions. First, we build a game-play benchmark that jointly measures instruction following and functional capabilities in 30 languages: the 24 official EU languages, plus six more chosen for comparison and anchoring. Second, we evaluate nine open-weight and commercial LLMs and show that the gap between high- and low-resource languages is large under interactive use, and that it falls almost entirely on the open-weight models. Third, we relate performance to training-data availability, tokeniser fertility and the economic weight of a language\footnote{Source code: \url{https://github.com/clembench/multilingual}\\Leaderboard of LLMs: \url{https://clembench.github.io/leaderboard.html}}.

% Related Work section. Comparison table is Table~\ref{tab:rw-comparison} (full-width, two-column span).
% Symbol macros (guarded so they can move to the preamble later without clashes)

\section{Related Work}
\label{sec:related}

% Figure 1 is defined here (not in Sec 3) so the float is encountered on page 2
% and can be placed at the top of page 3, next to the methodology text.
\begin{figure*}[t]
    \centering
    \includegraphics[width=1\linewidth]{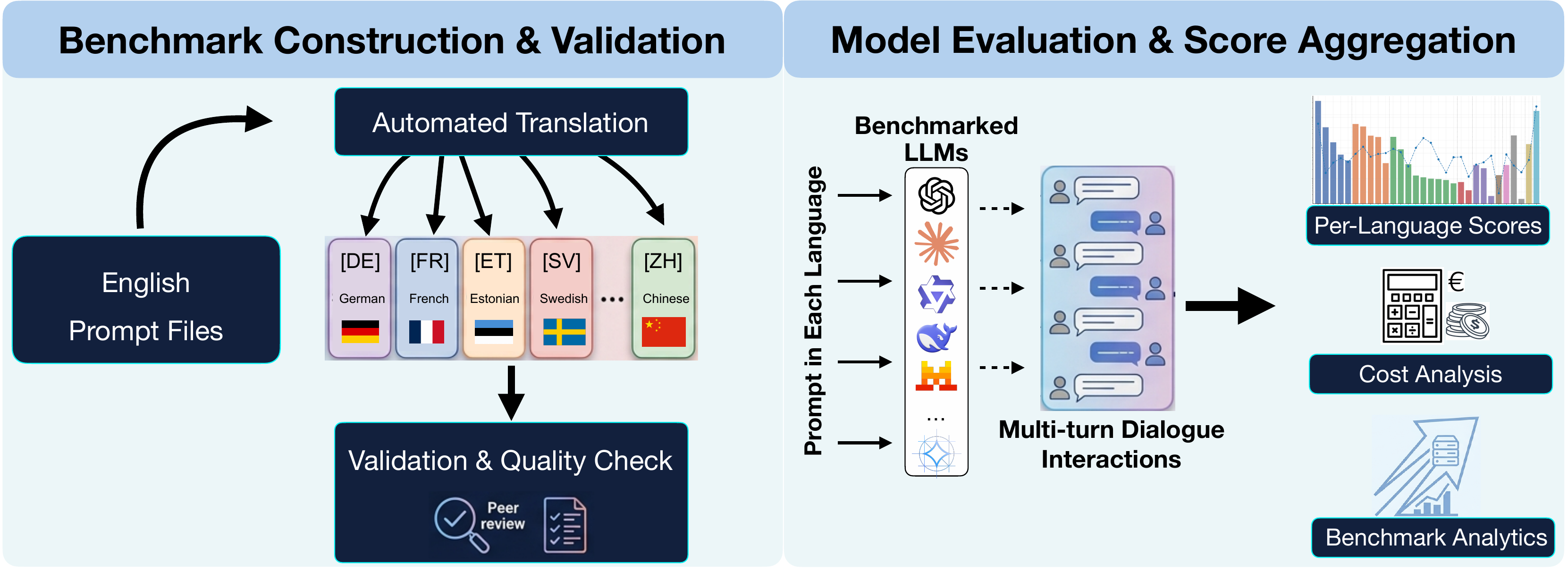}
    \caption{Overview of the multilingual benchmark construction and score aggregation across evaluated LLMs.}
    \label{fig:pipeline-overview}
\end{figure*}

\paragraph{Single-turn knowledge benchmarks.}
The dominant paradigm machine-translates English test sets, most prominently MMLU \citep{DBLP:conf/iclr/HendrycksBBZMSS21}, into other languages. Global MMLU \citep{DBLP:conf/acl/SinghRFANVLMLSN25}, MMLU-ProX \citep{DBLP:conf/emnlp/XuanYQZXFLXWGLJLLYDGLXJ25} and MuBench \citep{DBLP:journals/corr/abs-2506-19468} follow this recipe, as do PolyMath \citep{DBLP:journals/corr/abs-2504-18428} and OneRuler \citep{DBLP:journals/corr/abs-2503-01996}. Such scores conflate target-language ability with translation quality and stay Anglo-centric \citep{DBLP:conf/acl/SinghRFANVLMLSN25}. Native-source benchmarks such as M3Exam \citep{DBLP:conf/nips/ZhangAGCB23}, INCLUDE \citep{DBLP:conf/iclr/RomanouFSNSMACH25} and MultiLoKo \citep{DBLP:journals/corr/abs-2504-10356} arose as a corrective, but remain static, single-turn and contamination-prone, measuring knowledge recall rather than language use.

\paragraph{Multi-task suites and frameworks.}
A second line widens coverage by aggregating many such tasks \citep{DBLP:conf/emnlp/AhujaDHORJNGSAB23,DBLP:conf/naacl/AhujaAGWSOHJABS24,DBLP:conf/emnlp/LaiNVMDBN23,DBLP:conf/acl/UstunAYKDOBSOKV24,nielsen-2023-scandeval}. BenchMAX \citep{DBLP:conf/emnlp/HuangZHHLHY25} is closest in spirit, but probes each capability in a separate static subtask, so its scores cannot show how capabilities interact.

\paragraph{Instruction following and multi-turn evaluation.}
Multi-IF \citep{DBLP:journals/corr/abs-2410-15553} and XIFBench \citep{DBLP:journals/corr/abs-2503-07539} evaluate multilingual instruction following through verifiable constraints; Multi-IF adds turns, but they accumulate pre-scripted constraints rather than responding to the model's output, isolating compliance from what instructions serve.

\paragraph{Low-resource coverage and benchmark quality.}
Regional benchmarks document a persistent performance gap for low-resource languages in Africa \citep{DBLP:conf/naacl/AdelaniOAZAHOHBLCBSKMKY25,DBLP:conf/acl/OjoOOOLSA25,adebara-etal-2025-evaluating}, Southeast Asia \citep{lovenia-etal-2024-seacrowd}, India \citep{kakwani-etal-2020-indicnlpsuite} and the Turkic family \citep{senel-etal-2024-kardes,isbarov-etal-2025-tumlu}, as well as for morphologically rich languages \citep{DBLP:journals/corr/abs-2511-10664}. European-language evaluation likewise relies on translated English test sets \citep{DBLP:journals/corr/abs-2410-08928,DBLP:journals/corr/abs-2502-12895}. Yet such benchmarks correlate poorly with human judgments \citep{DBLP:journals/corr/abs-2504-15521}, and MCQ scores fluctuate with answer order and prompt \citep{DBLP:conf/naacl/AzimeTBCBAANYGtSSK25}.

\paragraph{Game-based evaluation and our approach.}
A parallel line of work evaluates LLMs as agents in interactive game environments, among them TextWorld \citep{textworld}, TextArena \citep{DBLP:journals/corr/abs-2504-11442}, TALES \citep{DBLP:journals/corr/abs-2504-14128} and GameArena \citep{DBLP:conf/iclr/HuLXJSJZ25}. These are developed and run in English. The clembench framework we build on has been instantiated in English, German and Italian \citep{DBLP:journals/corr/abs-2507-08491}. This shows that the approach ports across languages, but it leaves open whether it scales to a full institutional language set, and at what cost. Our benchmark goes beyond both lines (Table~\ref{tab:rw-comparison}): evaluation proceeds through goal-directed dialogue contingent on the model's own moves, with generated rather than frozen instances, scoring formal and functional competence in the same episode across 30 languages.

\section{Methodology}
\label{sec:methodology}

% Table with the games (tab:games) is defined here so the float is encountered on page 3
% and can be placed at the top of page 4, close to where it is discussed (Sec. 4.2).
\begin{table*}[t]
\centering
\small
\setlength{\tabcolsep}{4pt}
\begin{tabular}{p{3.6cm} l l}
\toprule
\textbf{Capability} & \textbf{Game} & \textbf{Description} \\
\midrule
Lexical \& world knowledge
  & Taboo        & Describe a target word without using the listed taboo words. \\
  & Codenames    & Give a one-word clue linking several words on the board. \\
  & Wordle       & Guess a hidden word from colour-coded letter feedback. \\
\midrule
Grounded reference
  & Reference Game & Describe a target among distractors for a partner to pick out. \\
  & Image Game     & Instruct a partner to recreate a pixel-art grid. \\
\midrule
Discourse \& grounding
  & GuessWhat         & Identify a target object through yes/no questions. \\
  & Match-It          & Decide in dialogue whether two ASCII images are identical. \\
  & Private \& Shared & Track which information is private vs.\ mutually known. \\
\midrule
Strategic \& social reasoning
  & Deal or No Deal & Negotiate an item split meeting private utility targets. \\
  & Hot Air Balloon & Agree on items to discard under a survival constraint. \\
  & Clean Up        & Collaboratively sort objects into correct locations. \\
\midrule
Spatial reasoning \& planning
  & TextMapWorld        & Navigate a text-based map to reach a goal location. \\
  & TextMapWorld (Graph) & Reason over the graph structure of a multi-room map. \\
  & TextMapWorld (Room)  & Plan fine-grained movement within individual rooms. \\
\bottomrule
\end{tabular}
\caption{The 14 games used in this benchmark, grouped by the capability cluster they primarily assess.}
\label{tab:games}
\end{table*}

\subsection{Dialogue Game-Based Evaluation}
\label{sec:clembench}

LLM evaluation is dominated by two paradigms \citep{DBLP:journals/corr/abs-2507-08491}. \emph{Reference-based} evaluation on static benchmarks offers control over what is tested, but it is single-turn, costly to extend, and prone to leakage and saturation. \emph{Preference-based} evaluation, such as LM Arena, tests real interactive use but offers no control and is not replicable. \citet{DBLP:journals/corr/abs-2507-08491} argue for a third paradigm that combines the strengths of both. \emph{Dialogue game-based} evaluation is controlled, repeatable, multi-turn and reference-free: success is determined by reaching a goal state under formal rules, not by comparison to gold answers.

We adopt this paradigm through clembench \citep{chalamalasetti2023clembench} and the 14 games listed in Table~\ref{tab:games}. A \emph{game master} orchestrates self-play episodes: it prompts one or more \emph{player} LLMs, validates their responses against formal rules, and scores rule compliance and task success programmatically. The game mechanics themselves are language-agnostic. A game scales to \emph{any} language for which the language-dependent artefacts exist, namely prompt text, response-parsing rules, feedback messages and, for some games, word lists. No reference answers need to be authored. We exploit this to extend clembench to 30 languages, acquiring the artefacts via machine translation with subsequent validation; Figure~\ref{fig:pipeline-overview} shows the resulting pipeline, from localisation to per-language score aggregation. This differs from the translate-then-benchmark approach criticised in Section~\ref{sec:related}. There the translation is the test item and its reference answer; here it is only the interface. Success remains behavioural and is verified against the formal rules of the game, not against a gold answer that translation could corrupt.

\subsection{Game Localisation Pipeline}
\label{sec:localisation}

Localising a game means translating three kinds of artefact: the prompt text, the response-parsing rules and the feedback messages. These must stay mutually consistent, since a loose prompt translation can break the corresponding regex. Localisation therefore proceeds in two stages, each carried out by a different model. First, GPT-5.2 translates each game's file set in a single pass, preserving formatting markers, placeholders, and regex syntax. Second, Claude Sonnet 4.5 compares originals against translations, checks that prompts and regexes still agree, and corrects errors. Splitting the stages across providers reduces the risk that systematic errors of one model go undetected, and neither model is among those we evaluate (Section~\ref{sec:models}).

\paragraph{Word lists.}
Four games additionally require language-specific word lists. For \textbf{Taboo} we extracted target and related words from ConceptNet~5.7 \citep{speer2017conceptnet}. For \textbf{Wordle} we took 5-letter noun lemmas from Universal Dependencies v2.17 treebanks \citep{nivre-etal-2020-universal}, ranked by corpus frequency, and built the set of valid guesses from Wikipedia dumps. For \textbf{Codenames} and \textbf{GuessWhat} we translated the English lists using the same pipeline. Details are in Appendix~\ref{app:wordlists}.

\subsection{Manual Verification}
\label{sec:manual-verification}

We verified six languages manually, chosen for typological and resource diversity: German, Russian, Chinese, Turkish, Greek and Hungarian. Native speakers inspected the translated files, checked that the parsing rules match the prompt wording, and confirmed the translations were fit to run. This was repeated while developing the pipeline, so each round of corrections propagated to all 30 languages; the changes required were minimal, concerning keyword choices and punctuation. Each LLM is then run in self-play for every language independently, with prompts, responses, parsing and scoring all operating in the target language.

\section{Experimental Setup}
\label{sec:setup}

\subsection{Languages}
\label{sec:languages}

The benchmark covers 30 languages (Table~\ref{tab:languages}): the 24 official EU languages plus six others. Turkish, Arabic, Russian and Ukrainian have large speaker populations, and Chinese allows comparing models developed in China. Serbian, written here in Latin script, is close enough to Croatian to test whether two closely related languages receive comparable support. The selection spans five families, Indo-European (Germanic, Romance, Slavic, Baltic, Celtic and Hellenic branches), Uralic, Turkic, Afro-Asiatic and Sino-Tibetan, and five scripts: Latin, Cyrillic, Greek, Arabic and Han.

\begin{table}[t]
\centering
\small
\setlength{\tabcolsep}{3pt}
\resizebox{\columnwidth}{!}{%
\begin{tabular}{ll ll ll}
\toprule
\textbf{Code} & \textbf{Language} & \textbf{Code} & \textbf{Language} & \textbf{Code} & \textbf{Language} \\
\midrule
ar & Arabic     & fr & French     & pt & Portuguese \\
bg & Bulgarian  & ga & Irish      & ro & Romanian \\
cs & Czech      & hr & Croatian   & ru & Russian \\
da & Danish     & hu & Hungarian  & sk & Slovak \\
de & German     & it & Italian    & sl & Slovenian \\
el & Greek      & lt & Lithuanian & sr & Serbian \\
en & English    & lv & Latvian    & sv & Swedish \\
es & Spanish    & mt & Maltese    & tr & Turkish \\
et & Estonian   & nl & Dutch      & uk & Ukrainian \\
fi & Finnish    & pl & Polish     & zh & Chinese \\
\bottomrule
\end{tabular}%
}
\caption{The 30 languages included in the benchmark.}
\label{tab:languages}
\end{table}

\subsection{Games and Capabilities}
\label{sec:games}

We include 14 dialogue games from the clembench suite, grouped in Table~\ref{tab:games} into five capability clusters, each targeting a facet of language use that static benchmarks cannot probe. Wordle is excluded for Chinese, because the game requires five-letter words whereas Chinese is character-based.

\subsection{Models}
\label{sec:models}

We evaluate nine LLMs, with details in Table~\ref{tab:model-sources}. Two are commercial, \textit{Claude Opus 4.8} and \textit{GPT-5.4}. Seven are recent multilingual open-weight models: \textit{GLM-5.2}, \textit{Mistral-Large-3}, \textit{Apertus-1.5-8B}, \textit{Gemma-4-26B}, \textit{Qwen3.6-35B}, \textit{DeepSeek-V4-Pro} and \textit{Nemotron-3-Ultra}. The commercial models were accessed through their provider APIs. GLM-5.2, DeepSeek-V4-Pro, Nemotron-3-Ultra and Mistral-Large-3 were accessed through OpenRouter, and Qwen3.6, Gemma-4 and Apertus-1.5 were run locally on NVIDIA A100 GPUs.

\paragraph{Hyperparameters.}
All models run in their default configuration with temperature $=1$ and no other decoding parameters set, so reasoning behaviour follows each provider's default. The output limit is 500 tokens per turn for all games except \textit{Clean Up} and \textit{Hot Air Balloon}, which use 2{,}000 tokens as their turns are substantially longer. Each language is evaluated in a single run using the same game instances for every model: 400 episodes per language per model (380 for Chinese), distributed across the 14 games as listed in Table~\ref{tab:episodes}, for a total of roughly 108{,}000 episodes.

\subsection{Metrics}
\label{sec:metrics}

Each episode is scored along two dimensions. \textbf{\%Played} is the share of episodes played to completion without aborting, capturing instruction-following and format compliance. \textbf{Quality} is the mean task-specific main score over non-aborted episodes, reflecting how well the model solved the task when it engaged. Both are aggregated into the \textbf{clemscore}, the normalised product of \%Played and Quality scaled to $[0, 100]$, so a high clemscore requires a model to play reliably and perform well.

\section{Results}
\label{sec:results}

\subsection{Performance across Languages}

\begin{table}[t!]
\centering
\scriptsize
\setlength{\tabcolsep}{2.5pt}
\renewcommand{\arraystretch}{1.05}
\begin{tabular}{@{}l *{9}{r} *{1}{r}@{}}
\toprule
\textbf{Lang}
  & \rotatebox{90}{\texttt{GPT-5.4}}
  & \rotatebox{90}{\texttt{Opus-4.8}}
  & \rotatebox{90}{\texttt{GLM-5.2}}
  & \rotatebox{90}{\texttt{DS-V4}}
  & \rotatebox{90}{\texttt{Nemotron}}
  & \rotatebox{90}{\texttt{Gemma4}}
  & \rotatebox{90}{\texttt{Qwen3.6}}
  & \rotatebox{90}{\texttt{Mistral-L3}}
  & \rotatebox{90}{\texttt{Apertus}}
  & \rotatebox{90}{\textbf{Avg}} \\
\midrule
ar & \cellcolor{teal!52}79.1 & \cellcolor{teal!52}69.1 & \cellcolor{teal!36}60.2 & \cellcolor{teal!28}52.1 & \cellcolor{teal!20}47.1 & \cellcolor{teal!20}49.9 & \cellcolor{teal!13}34.7 & \cellcolor{teal!13}39.3 & \cellcolor{teal!5}11.8 & \cellcolor{teal!20}49.3 \\
\rowcolor{gray!12} bg & \cellcolor{teal!60}92.4 & \cellcolor{teal!52}80.6 & \cellcolor{teal!52}68.9 & \cellcolor{teal!44}61.2 & \cellcolor{teal!28}52.3 & \cellcolor{teal!36}60.8 & \cellcolor{teal!28}52.4 & \cellcolor{teal!13}41.7 & \cellcolor{teal!5}17.6 & \cellcolor{teal!36}58.7 \\
cs & \cellcolor{teal!60}90.6 & \cellcolor{teal!52}81.7 & \cellcolor{teal!36}60.1 & \cellcolor{teal!36}58.6 & \cellcolor{teal!36}55.7 & \cellcolor{teal!28}52.8 & \cellcolor{teal!28}50.4 & \cellcolor{teal!13}41.8 & \cellcolor{teal!5}14.1 & \cellcolor{teal!36}56.2 \\
\rowcolor{gray!12} da & \cellcolor{teal!60}89.9 & \cellcolor{teal!60}83.0 & \cellcolor{teal!52}68.5 & \cellcolor{teal!44}65.0 & \cellcolor{teal!36}58.5 & \cellcolor{teal!36}59.2 & \cellcolor{teal!20}47.4 & \cellcolor{teal!20}44.8 & \cellcolor{teal!5}11.3 & \cellcolor{teal!36}58.6 \\
de & \cellcolor{teal!60}90.7 & \cellcolor{teal!60}85.5 & \cellcolor{teal!44}67.3 & \cellcolor{teal!52}70.1 & \cellcolor{teal!44}63.0 & \cellcolor{teal!36}57.9 & \cellcolor{teal!13}42.5 & \cellcolor{teal!20}49.4 & \cellcolor{teal!5}16.2 & \cellcolor{teal!36}60.3 \\
\rowcolor{gray!12} el & \cellcolor{teal!60}91.4 & \cellcolor{teal!52}75.5 & \cellcolor{teal!52}69.9 & \cellcolor{teal!44}63.1 & \cellcolor{teal!28}51.3 & \cellcolor{teal!20}47.6 & \cellcolor{teal!28}54.9 & \cellcolor{teal!13}34.7 & \cellcolor{teal!5}19.1 & \cellcolor{teal!36}56.4 \\
en & \cellcolor{teal!60}91.8 & \cellcolor{teal!60}84.4 & \cellcolor{teal!44}65.6 & \cellcolor{teal!44}64.8 & \cellcolor{teal!52}68.2 & \cellcolor{teal!28}50.6 & \cellcolor{teal!28}55.6 & \cellcolor{teal!28}54.6 & \cellcolor{teal!5}21.4 & \cellcolor{teal!44}61.9 \\
\rowcolor{gray!12} es & \cellcolor{teal!60}85.2 & \cellcolor{teal!52}70.0 & \cellcolor{teal!44}62.0 & \cellcolor{teal!36}57.8 & \cellcolor{teal!36}56.2 & \cellcolor{teal!28}52.9 & \cellcolor{teal!20}44.4 & \cellcolor{teal!20}45.1 & \cellcolor{teal!5}17.0 & \cellcolor{teal!28}54.5 \\
et & \cellcolor{teal!60}92.3 & \cellcolor{teal!52}81.8 & \cellcolor{teal!44}65.5 & \cellcolor{teal!36}58.1 & \cellcolor{teal!13}39.8 & \cellcolor{teal!13}41.4 & \cellcolor{teal!20}46.8 & \cellcolor{teal!13}31.5 & \cellcolor{teal!5}8.2 & \cellcolor{teal!28}51.7 \\
\rowcolor{gray!12} fi & \cellcolor{teal!60}93.6 & \cellcolor{teal!52}79.6 & \cellcolor{teal!44}61.1 & \cellcolor{teal!44}61.6 & \cellcolor{teal!28}54.8 & \cellcolor{teal!28}50.6 & \cellcolor{teal!13}42.4 & \cellcolor{teal!20}43.4 & \cellcolor{teal!5}18.6 & \cellcolor{teal!36}56.2 \\
fr & \cellcolor{teal!60}92.3 & \cellcolor{teal!60}83.5 & \cellcolor{teal!52}67.4 & \cellcolor{teal!28}52.4 & \cellcolor{teal!36}58.0 & \cellcolor{teal!28}52.9 & \cellcolor{teal!28}53.9 & \cellcolor{teal!20}45.8 & \cellcolor{teal!5}20.3 & \cellcolor{teal!36}58.5 \\
\rowcolor{gray!12} ga & \cellcolor{teal!60}84.9 & \cellcolor{teal!52}76.3 & \cellcolor{teal!13}40.5 & \cellcolor{teal!13}37.8 & \cellcolor{teal!13}30.4 & \cellcolor{teal!5}16.9 & \cellcolor{teal!13}32.4 & \cellcolor{teal!5}19.3 & \cellcolor{teal!5}6.9 & \cellcolor{teal!13}38.4 \\
hr & \cellcolor{teal!60}93.0 & \cellcolor{teal!60}83.7 & \cellcolor{teal!44}64.2 & \cellcolor{teal!36}56.4 & \cellcolor{teal!28}53.2 & \cellcolor{teal!28}55.2 & \cellcolor{teal!20}46.0 & \cellcolor{teal!13}42.1 & \cellcolor{teal!5}18.7 & \cellcolor{teal!36}56.9 \\
\rowcolor{gray!12} hu & \cellcolor{teal!52}81.1 & \cellcolor{teal!52}78.3 & \cellcolor{teal!44}63.4 & \cellcolor{teal!28}52.6 & \cellcolor{teal!20}46.2 & \cellcolor{teal!20}49.6 & \cellcolor{teal!13}40.0 & \cellcolor{teal!13}40.1 & \cellcolor{teal!5}17.6 & \cellcolor{teal!28}52.1 \\
it & \cellcolor{teal!60}93.9 & \cellcolor{teal!52}80.9 & \cellcolor{teal!44}66.8 & \cellcolor{teal!44}65.7 & \cellcolor{teal!36}61.0 & \cellcolor{teal!36}59.6 & \cellcolor{teal!20}46.8 & \cellcolor{teal!13}37.1 & \cellcolor{teal!5}14.6 & \cellcolor{teal!36}58.5 \\
\rowcolor{gray!12} lt & \cellcolor{teal!52}77.9 & \cellcolor{teal!52}74.1 & \cellcolor{teal!36}59.5 & \cellcolor{teal!13}40.3 & \cellcolor{teal!20}46.2 & \cellcolor{teal!13}37.0 & \cellcolor{teal!13}42.4 & \cellcolor{teal!5}25.7 & \cellcolor{teal!5}16.2 & \cellcolor{teal!20}46.6 \\
lv & \cellcolor{teal!60}91.8 & \cellcolor{teal!52}79.9 & \cellcolor{teal!44}61.9 & \cellcolor{teal!36}59.5 & \cellcolor{teal!20}45.4 & \cellcolor{teal!20}49.0 & \cellcolor{teal!28}50.2 & \cellcolor{teal!13}32.7 & \cellcolor{teal!5}14.3 & \cellcolor{teal!28}53.9 \\
\rowcolor{gray!12} mt & \cellcolor{teal!60}92.0 & \cellcolor{teal!52}75.6 & \cellcolor{teal!20}45.4 & \cellcolor{teal!44}64.3 & \cellcolor{teal!13}32.7 & \cellcolor{teal!13}34.2 & \cellcolor{teal!13}40.1 & \cellcolor{teal!5}12.3 & \cellcolor{teal!5}2.3 & \cellcolor{teal!20}44.3 \\
nl & \cellcolor{teal!60}90.7 & \cellcolor{teal!60}83.5 & \cellcolor{teal!44}66.8 & \cellcolor{teal!44}63.9 & \cellcolor{teal!44}62.7 & \cellcolor{teal!36}60.2 & \cellcolor{teal!20}49.5 & \cellcolor{teal!28}54.3 & \cellcolor{teal!5}15.0 & \cellcolor{teal!36}60.7 \\
\rowcolor{gray!12} pl & \cellcolor{teal!60}89.5 & \cellcolor{teal!60}82.3 & \cellcolor{teal!44}64.3 & \cellcolor{teal!44}62.5 & \cellcolor{teal!36}57.1 & \cellcolor{teal!36}59.7 & \cellcolor{teal!20}45.9 & \cellcolor{teal!20}49.6 & \cellcolor{teal!5}22.0 & \cellcolor{teal!36}59.2 \\
pt & \cellcolor{teal!60}85.9 & \cellcolor{teal!52}71.2 & \cellcolor{teal!36}56.1 & \cellcolor{teal!28}53.5 & \cellcolor{teal!36}58.1 & \cellcolor{teal!36}55.9 & \cellcolor{teal!20}49.3 & \cellcolor{teal!13}41.8 & \cellcolor{teal!5}19.0 & \cellcolor{teal!28}54.5 \\
\rowcolor{gray!12} ro & \cellcolor{teal!52}74.1 & \cellcolor{teal!52}73.0 & \cellcolor{teal!44}61.9 & \cellcolor{teal!28}50.3 & \cellcolor{teal!28}53.9 & \cellcolor{teal!20}43.0 & \cellcolor{teal!20}45.7 & \cellcolor{teal!13}36.0 & \cellcolor{teal!5}17.8 & \cellcolor{teal!28}50.6 \\
ru & \cellcolor{teal!60}89.3 & \cellcolor{teal!52}80.1 & \cellcolor{teal!44}62.3 & \cellcolor{teal!44}62.5 & \cellcolor{teal!36}59.0 & \cellcolor{teal!36}58.0 & \cellcolor{teal!28}53.8 & \cellcolor{teal!20}46.8 & \cellcolor{teal!5}23.0 & \cellcolor{teal!36}59.4 \\
\rowcolor{gray!12} sk & \cellcolor{teal!60}90.3 & \cellcolor{teal!52}82.1 & \cellcolor{teal!36}59.4 & \cellcolor{teal!36}59.3 & \cellcolor{teal!20}45.5 & \cellcolor{teal!28}53.6 & \cellcolor{teal!20}48.6 & \cellcolor{teal!13}38.5 & \cellcolor{teal!5}12.9 & \cellcolor{teal!28}54.5 \\
sl & \cellcolor{teal!60}86.1 & \cellcolor{teal!52}78.7 & \cellcolor{teal!36}59.8 & \cellcolor{teal!28}55.1 & \cellcolor{teal!28}53.4 & \cellcolor{teal!13}36.6 & \cellcolor{teal!20}48.8 & \cellcolor{teal!20}43.9 & \cellcolor{teal!5}18.8 & \cellcolor{teal!28}53.5 \\
\rowcolor{gray!12} sr & \cellcolor{teal!60}90.6 & \cellcolor{teal!60}84.6 & \cellcolor{teal!36}60.7 & \cellcolor{teal!36}58.7 & \cellcolor{teal!20}48.6 & \cellcolor{teal!36}60.2 & \cellcolor{teal!13}42.7 & \cellcolor{teal!13}37.0 & \cellcolor{teal!5}19.6 & \cellcolor{teal!36}55.8 \\
sv & \cellcolor{teal!60}87.8 & \cellcolor{teal!52}80.6 & \cellcolor{teal!44}65.2 & \cellcolor{teal!44}65.5 & \cellcolor{teal!44}61.2 & \cellcolor{teal!36}56.8 & \cellcolor{teal!20}45.9 & \cellcolor{teal!28}51.3 & \cellcolor{teal!5}20.5 & \cellcolor{teal!36}59.4 \\
\rowcolor{gray!12} tr & \cellcolor{teal!60}91.0 & \cellcolor{teal!52}75.0 & \cellcolor{teal!44}63.3 & \cellcolor{teal!44}63.5 & \cellcolor{teal!28}51.4 & \cellcolor{teal!13}42.0 & \cellcolor{teal!13}41.4 & \cellcolor{teal!13}33.2 & \cellcolor{teal!5}9.4 & \cellcolor{teal!28}52.3 \\
uk & \cellcolor{teal!60}93.0 & \cellcolor{teal!52}79.6 & \cellcolor{teal!44}67.0 & \cellcolor{teal!36}55.7 & \cellcolor{teal!28}50.4 & \cellcolor{teal!36}56.7 & \cellcolor{teal!28}50.0 & \cellcolor{teal!28}50.4 & \cellcolor{teal!5}17.7 & \cellcolor{teal!36}57.8 \\
\rowcolor{gray!12} zh & \cellcolor{teal!60}91.3 & \cellcolor{teal!52}73.9 & \cellcolor{teal!52}71.3 & \cellcolor{teal!44}62.8 & \cellcolor{teal!52}68.4 & \cellcolor{teal!20}49.0 & \cellcolor{teal!44}61.7 & \cellcolor{teal!28}50.7 & \cellcolor{teal!5}18.0 & \cellcolor{teal!36}60.8 \\
\midrule
Avg & \cellcolor{teal!60}88.8 & \cellcolor{teal!52}78.9 & \cellcolor{teal!44}62.6 & \cellcolor{teal!36}58.5 & \cellcolor{teal!28}53.0 & \cellcolor{teal!28}50.3 & \cellcolor{teal!20}46.9 & \cellcolor{teal!13}40.5 & \cellcolor{teal!5}16.0 & \cellcolor{teal!28}55.1 \\
STD & \cellcolor{teal!5}4.9 & \cellcolor{teal!5}4.5 & \cellcolor{teal!5}6.3 & \cellcolor{teal!5}7.0 & \cellcolor{teal!5}8.8 & \cellcolor{teal!5}9.7 & \cellcolor{teal!5}6.1 & \cellcolor{teal!5}9.6 & \cellcolor{teal!5}4.7 & \cellcolor{teal!5}21.2 \\
\bottomrule
\end{tabular}%
\caption{Overall clemscore per language (rows) and model (columns): mean \% Played across the 14 games times mean Quality Score across the 14 games, ordered by mean clemscore. Cells are shaded by octile of the clemscore distribution (darker = higher); the best model per language is in \textbf{bold}.}
\label{tab:overall-results}
\end{table}

\paragraph{Overall Comparison.}
Table~\ref{tab:overall-results} reports \textit{clemscore} for all models across the 30 languages (\%Played and Quality separately in Tables~\ref{tab:played-per-language} and~\ref{tab:quality-per-language}). The dominant pattern is a wide gap between the two commercial models and everything else: \textit{GPT-5.4} and \textit{Claude Opus 4.8} lead the best open-weight model, \textit{GLM-5.2}, by 26 and 16 points on average. In every one of the 24 official EU languages, including Irish, Maltese and Latvian, both commercial models score above \emph{every} open-weight model, by margins running from 5.6 points in Greek to 35.8 in Irish. Across the EU-24 the commercial systems bottom out at 74.1 (GPT-5.4, Romanian) and 70.0 (Opus~4.8, Spanish), while the highest open-weight result anywhere in the benchmark is 71.3 (GLM-5.2, Chinese). Against their own English results, the commercial systems return 80.7\% and 83.0\% in their weakest EU language, whereas the open-weight models return between 61.7\% and 10.7\%, so their deficit is a collapse concentrated in specific languages rather than uniform weakness. For all seven that weakest language is Irish or Maltese. The closely related Serbian and Croatian (Section~\ref{sec:languages}) do receive comparable support, at 55.8 and 56.9 averaged over models. The material itself is playable throughout: in each of the 30 languages at least one model completes 86\% or more of its episodes. Parameter count does not explain the ordering. The 675B \textit{Mistral-Large-3} averages 40.5, below the far smaller \textit{Gemma-4} and \textit{Qwen3.6}, so coverage of the EU-24 is not simply a matter of scale. \textbf{The strongest open-weight model, GLM-5.2, spans 69.9 (Greek) to 40.5 (Irish) across the EU-24; none of the seven covers the full set.}

Provenance helps, but it does not confer advantage. Chinese is the single best language of all 30 for both \textit{Qwen3.6} (61.7) and \textit{GLM-5.2} (71.3), while the third Chinese-developed model, \textit{DeepSeek-V4-Pro}, scores highest in German and places Chinese tenth. Yet the best Chinese result of any model in the benchmark belongs to \textit{GPT-5.4}, at 91.3 --- some 20 points above the strongest Chinese-developed system in its own language. Being trained where a language is spoken lifts a model within its own range without closing the gap to the commercial systems. Averaged over models, Chinese ranks second of the 30 behind English, consistent with its third place in available web text, though its 13-game average flatters it by two to six points relative to languages scored over all 14 (Appendix~\ref{app:episodes}).

\begin{figure}[t!]
    \centering
    \includegraphics[width=1.0\linewidth]{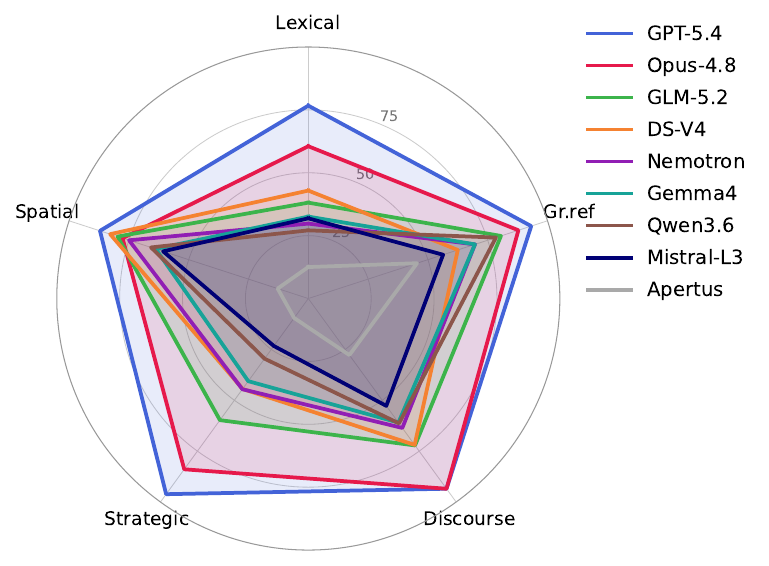}
    \caption{Model performance averages across languages for each language capability (game results are averaged)}
    \label{fig:radar_all_capabilities}
\end{figure}

Averaging over languages, Figure~\ref{fig:radar_all_capabilities} breaks performance down by capability cluster: lexical knowledge and strategic reasoning are hardest, at 39.1 and 48.0, while grounded reference and spatial reasoning are the strongest, at 71.5 and 67.3. Spatial reasoning is also the cluster that separates the open-weight models most sharply, spanning 70 points between the best and the weakest of the seven (per-language detail in Tables~\ref{tab:cgmT-lexical}--\ref{tab:cgmT-spatial}).

\begin{figure*}[t!]
    \centering
    \includegraphics[width=1.0\linewidth]{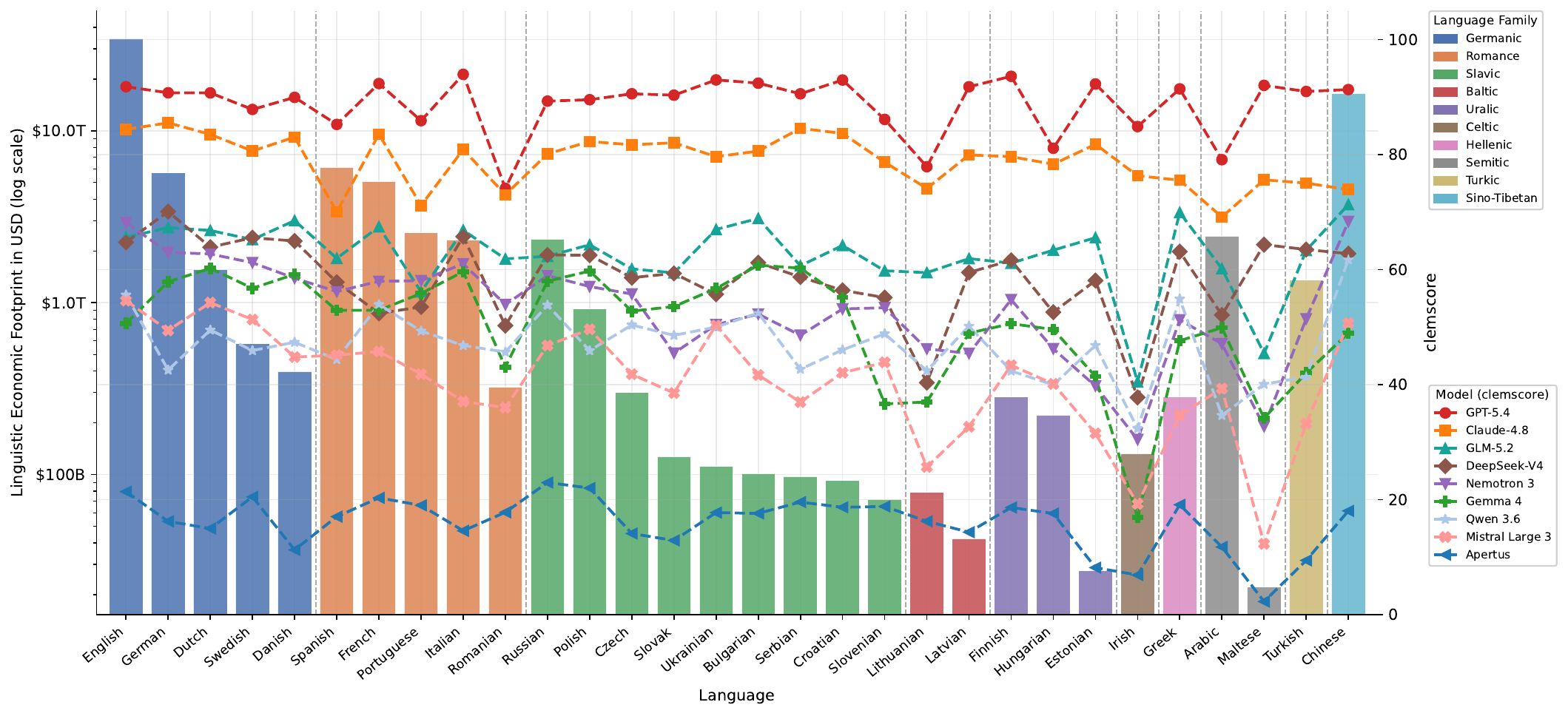}
    \caption{LEF (bars, left axis, log scale) against clemscore (lines, right axis) per language, grouped by family. Open-weight models drop for the smaller-footprint languages, while the commercial systems show no consistent trend, varying over 74.1--93.9 and 69.1--85.5.}
    \label{fig:lef}
\end{figure*}

\paragraph{Correlation with other Benchmarks.}
To situate \textit{clemscore} among established evaluations, we compare the English \textit{clemscore} ranking of five frontier models against their rankings on LMArena\footnote{\url{https://lmarena.ai/leaderboard/text/overall}}, GPQA Diamond~\cite{DBLP:journals/corr/abs-2311-12022}, HLE~\cite{DBLP:journals/corr/abs-2501-14249} and the agentic BrowseComp~\cite{DBLP:journals/corr/abs-2504-12516}, using Kendall's~$\tau$. The five models are GPT-5.4, Claude Opus 4.8, Nemotron-3-Ultra, GLM-5.2 and DeepSeek-V4-Pro. We restrict the comparison to English, as those benchmarks cover no other language for all five.
Agreement is moderate and decreases as the target capability moves away from interactive language use: $\tau = 0.60$ against LMArena human preferences, $0.40$ against GPQA Diamond, $0.20$ against HLE, and $0.00$ against BrowseComp.
Much of the disagreement comes from a single model: \textit{Nemotron-3-Ultra} ranks third in English \textit{clemscore} but 91st on LMArena, playing the games reliably without being a model users prefer in open-ended chat.
These external scores are vendor-reported under differing protocols, so they indicate approximate standing only. With $n = 5$ these coefficients are descriptive rather than significant, but the ordering is what we would expect if \textbf{dialogue-game evaluation measures something existing benchmarks do not capture}, rather than reproducing them. Full scores and ranking charts are in Appendix~\ref{sec:rank-correlation}.

\subsection{Economic Footprint, Data, and Tokenisation}
\label{sec:lef}

\begin{figure*}[t!]
    \centering
    \includegraphics[width=1.0\linewidth]{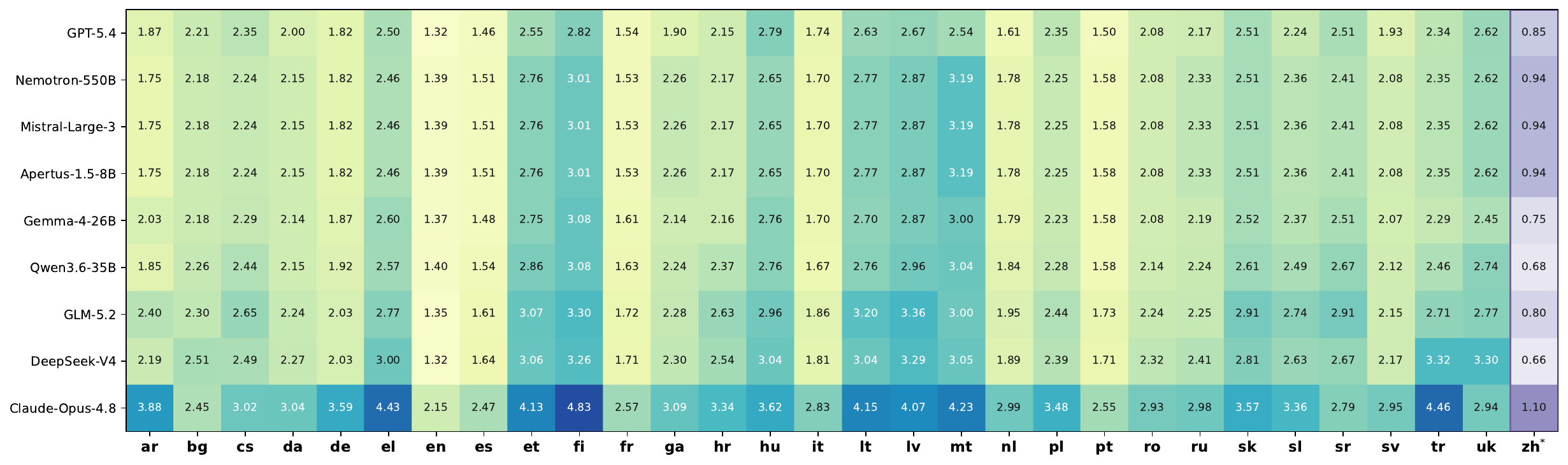}
    \caption{Token fertility per model and language. Chinese is character-based and not directly comparable.}
    \label{fig:token_fertility}
\end{figure*}

\paragraph{Linguistic Economic Footprint.} To relate performance to the real-world economic value of a language, we introduce the Linguistic Economic Footprint (LEF). We take each country's nominal GDP, weight it by the share of the population that speaks the language, and sum over the countries where the language holds (co-)official status:
\[
  \mathrm{LEF}(\ell) = \sum_{c \,\in\, \mathcal{C}_\ell} \mathrm{GDP}(c) \times \frac{\mathrm{speakers}(\ell,c)}{\mathrm{population}(c)}
\]
Prior work weighted demand mainly by speaker population \citep{blasi-etal-2022-systematic}; LEF makes the economic dimension explicit. Sources are given in Appendix~\ref{app:lef}.

Figure~\ref{fig:lef} plots \textit{LEF} against \textit{clemscore}. For the open-weight models, performance broadly follows a language's economic footprint: the Spearman correlation is positive for all seven, from $\rho = 0.22$ to $0.78$, though only the two strongest are significant at $n = 30$ (Figure~\ref{fig:lef-corr}). The two commercial systems break this pattern, at $\rho = -0.13$ and $-0.02$. \textit{GPT-5.4} scores as well on the three smallest footprints in our set (Maltese 92.0, Estonian 92.3, Latvian 91.8) as on English (91.8), and \textit{Claude Opus 4.8} drops by 8.8 points from English to Maltese (84.4 to 75.6), while remaining ahead of every open-weight model in each of those languages.

The same pattern holds for the volume of crawled text available per language, which spans four orders of magnitude across our set. It correlates with open-weight performance ($\rho = 0.72$, $p < 0.001$) but not with the commercial systems ($\rho = 0.03$ and $0.06$, both n.s.). Appendix~\ref{app:data-availability} gives the per-language figures. The two explanations are not independent, since richer economies also produce more web text, and neither fully determines performance: Serbian scores above Spanish (55.8 against 54.5) on a seventieth of the crawled text, and Croatian above Portuguese (56.9 against 54.5) on a tenth. LEF adds the demand side, spanning more than three orders of magnitude within our language set. If provider investment tracks market size, it alone will not close the gap for the smallest official languages, and the models that do serve them well are the closed ones. \textbf{Equal linguistic standing therefore depends on effort that market incentives do not reward, which is where public funding and open, language-targeted resources matter most.}

\begin{figure*}[t!]
    \centering
    \includegraphics[width=1.0\linewidth]{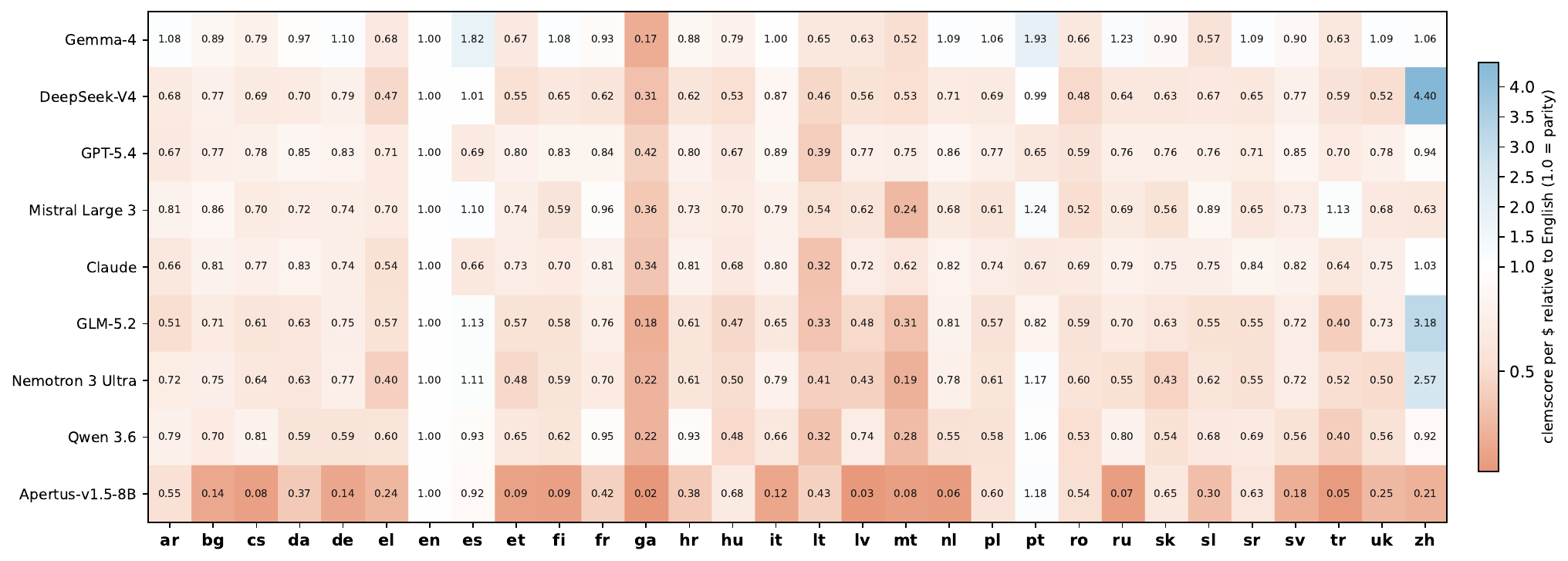}
    \caption{Value---clemscore points per dollar---normalised against the same model's English value: 1.0 matches what an English user of that model gets, lower values mean the same money buys less.}
    \label{fig:value-heatmap}
\end{figure*}

\paragraph{Token Fertility.}
Figure~\ref{fig:token_fertility} reports each model's \emph{token fertility}, the average number of tokens per word, which measures how well a tokeniser's vocabulary covers a language. The $\approx$1.4 observed for English with almost every model means words are mostly kept intact; the $\approx$3.0 for Maltese means every word is split into three pieces. The same content then occupies roughly twice the context, and since APIs bill per token, this translates directly into higher cost \citep{ahia-etal-2023-languages,DBLP:conf/nips/PetrovMTB23}.

Within the EU-24 every provider shows the same gradient: English, Spanish, Portuguese, French, Italian, Dutch and German sit at 1.4--1.9 tokens per word, while Finnish, Maltese, Latvian, Lithuanian, Estonian, Hungarian and Greek run at 2.6--3.1. Fertility is a downstream symptom rather than a cause: subword vocabularies are fitted to the text available, so languages with little of it get fewer dedicated units \citep{rust-etal-2021-good,velayuthan-sarveswaran-2025-egalitarian}. The profiles are near-identical across providers, which indicates that \textbf{no provider currently invests in vocabulary allocation for all EU-24 languages}; three of the nine are identical by construction, sharing Mistral's Tekken tokeniser (Appendix~\ref{app:fertility}). \textit{Claude Opus 4.8} is the one model that departs from this profile: it averages roughly 50\% more tokens per word than the median (Appendix~\ref{app:fertility}). \textbf{The tokeniser prices languages before a single forward-pass is made: speakers of Finnish, Maltese or Estonian spend roughly twice as many tokens, and therefore roughly twice the money and context, as English speakers for the same task.}

\subsection{Cost vs. Performance}

Strong absolute performance does not imply a language is served \emph{efficiently}, so we express results as \emph{value}: \textit{clemscore} points per dollar, normalised against the same model's English value (Figure~\ref{fig:value-heatmap}). Costs come from the token counts in the interaction transcripts and each model's list price (Table~\ref{tab:model-sources}); the per-language breakdown is in Table~\ref{tab:cost-per-language}.

The commercial systems, despite leading on raw scores, deliver markedly less value outside English: all 29 non-English languages cost \textit{GPT-5.4} more than English, and 28 of 29 cost \textit{Claude Opus 4.8} more, Chinese being the exception. In high-resource languages the premium is moderate and performance close to parity, so value stays near the English level. In the lower-resourced languages users \emph{pay more and get less}: the median model pays $2.2\times$ the English cost for Irish while returning $58\%$ of its English score. The mean, $3.6\times$, is inflated by \textit{Apertus}, whose Irish run costs $15\times$ its English one. \textit{Gemma-4} is the exception, with better-than-English value in twelve languages, led by Portuguese ($1.93$) and Spanish ($1.82$).

Pooled over all models and non-English languages, the median language costs $31\%$ more to run than English while scoring $10\%$ lower. Higher fertility inflates the billed token count, and lower performance reduces what those tokens buy. \textbf{Speakers of the smaller EU languages are thus charged a double premium: equal access to the same model, at the same price, still yields unequal service.}

\section{Conclusion}

We presented an interactive benchmark evaluating LLMs as agents in goal-directed dialogue games across 30 languages: the 24 official EU languages plus six others. Built by localising game files rather than translating a static test set, it is cheap to extend and to re-run as models evolve. No open-weight model we evaluated covers the EU-24: in every official language both commercial systems outscore every open-weight model, and the open-weight models pass 70 points in only two cells between them. That the commercial systems stay strong across the whole set cuts two ways. It shows linguistic parity is achievable: Irish and Maltese are not intrinsically harder, only under-served. Yet parity cannot come from public crawls, which offer four orders of magnitude less text for these languages. The models that serve Europe's smaller languages well are closed and non-European: equality means either accepting that dependence or building resources beyond the open web, such as public broadcast archives.

%digital sovereignty means either paying external providers a premium or investing in resources beyond the open web, such as public broadcast archives.

\section*{Limitations}

Our selection centres on the 24 official EU languages plus six others, so most languages of Africa, South and Southeast Asia and East Asia are absent; extending to them requires only the localised game files, which makes broader coverage a question of funding rather than of method. The open-weight set spans 8B to 675B parameters, so the open-versus-closed contrast is partly confounded with scale, though not entirely, since the 675B \textit{Mistral-Large-3} trails much smaller open-weight models. We evaluate one recent variant per family and omit the EU-funded models such as EuroLLM, Teuken and Salamandra, which would speak most directly to the policy framing of Section~\ref{sec:lef}; budget and run time, at two to four days per model across 30 languages, did not allow more. We consider those models the most useful addition to a future version.

Each language is run once, at temperature 1 and without a fixed seed, so we report no variance estimate and small differences between adjacent cells should not be read as meaningful. We also have no per-language human or random baseline, so scores are interpreted relative to other models rather than against an absolute difficulty scale. Games contribute unequally: Wordle depends on a five-letter constraint that is natural in some orthographies and marginal in others, and is dropped for Chinese, which raises Chinese scores by two to six points relative to languages scored over all 14 games. Only six languages were verified by native speakers, and a small number of game--language pairs still show near-zero completion for reasons we attribute to parsing rather than to competence.

Finally, reference-free scoring removes the gold answers that static benchmarks leak, but does not make the benchmark contamination-proof. Universal Dependencies, Wikipedia and ConceptNet are public, as is clembench with its published English results, so a model may have seen individual target words or transcripts of earlier runs. What cannot leak is the ability to play: a memorised Wordle target does not help a model that fails to follow the feedback protocol over five turns.

\section*{Ethics Statement}

The evaluated models can in principle produce inaccurate, biased, or unsafe text. In our setting this risk is limited. Outputs are generated only within tightly constrained dialogue games and serve only as input to the functions that score the interactions. They are not used for training, are not shown to end users, and are released only as game transcripts for reproducibility. The texts used for the token-fertility computation come from Wikipedia and are available under a free licence. No personal or sensitive data is collected, and no human subjects were involved beyond the verification of the translated game files.

% \section*{Acknowledgements}

\bibliography{custom,anthology_0,anthology_1}

\appendix

% Tighter float handling for the appendix only: the tables here are large and
% mostly self-contained, so allow denser pages and less space around floats.
\setlength{\textfloatsep}{8pt plus 2pt minus 2pt}
\setlength{\floatsep}{8pt plus 2pt minus 2pt}
\setlength{\intextsep}{8pt plus 2pt minus 2pt}
\setlength{\dbltextfloatsep}{8pt plus 2pt minus 2pt}
\setlength{\dblfloatsep}{8pt plus 2pt minus 2pt}
\setlength{\abovecaptionskip}{4pt}
\setlength{\belowcaptionskip}{0pt}
\renewcommand{\topfraction}{0.95}
\renewcommand{\bottomfraction}{0.95}
\renewcommand{\textfraction}{0.05}
\renewcommand{\floatpagefraction}{0.80}
\renewcommand{\dbltopfraction}{0.95}
\renewcommand{\dblfloatpagefraction}{0.80}

% Let long model URLs break inside narrow table columns.
\makeatletter
\g@addto@macro\UrlBreaks{%
  \do\A\do\B\do\C\do\D\do\E\do\F\do\G\do\H\do\I\do\J\do\K\do\L\do\M%
  \do\N\do\O\do\P\do\Q\do\R\do\S\do\T\do\U\do\V\do\W\do\X\do\Y\do\Z%
  \do\0\do\1\do\2\do\3\do\4\do\5\do\6\do\7\do\8\do\9}
\makeatother

\section{Word List Construction}
\label{app:wordlists}

Four games require per-language word lists in addition to translated prompts: Taboo, Wordle, Codenames and GuessWhat.

\paragraph{Taboo.}
Target words and their taboo (forbidden) words come from ConceptNet~5.7 \citep{speer2017conceptnet}, a multilingual commonsense knowledge graph. We iterate over all $\sim$36M assertions and keep only \texttt{/r/RelatedTo} edges with relation weight $\geq 2.0$, which filters for semantically close noun pairs with high crowdsourced confidence, and use \texttt{/r/FormOf} relations to map surface forms to their base forms so that inflectional variants do not appear separately. For each candidate target we then take up to three related words by descending weight, discarding targets with fewer than three qualifying neighbours, and retain up to 1{,}000 target--taboo pairs per language. A final pass removes potentially offensive or ambiguous entries using a manually curated blocklist.

\paragraph{Wordle.}
Wordle needs two lists per language: the hidden \emph{target words} and a larger set of \emph{allowed guesses}. Targets come from Universal Dependencies v2.17 \citep{nivre-etal-2020-universal}, which provides gold part-of-speech tags and lemmas for over 120 languages. For each language we take all tokens tagged \texttt{NOUN}, keep the lemmas of exactly five characters, and rank them by corpus frequency; the 100 most frequent form the \emph{easy} target list and the remainder the \emph{medium} list. Allowed guesses come from Wikipedia dumps: after lowercasing and removing punctuation, all five-character tokens are retained, which covers inflected forms and proper nouns beyond the UD-derived targets.

\paragraph{Codenames and GuessWhat.}
Both use closed lists whose playability depends on the words being common, unambiguous nouns, which automatic extraction cannot guarantee. We therefore translated the English lists with the pipeline of Section~\ref{sec:localisation} (GPT-5.2 translating, Claude Sonnet 4.5 validating). Spot-checks on the manually verified languages (Section~\ref{sec:manual-verification}) confirmed that the translations keep the intended register and avoid culturally inappropriate items.

\section{Episodes per Game}
\label{app:episodes}

Each model plays 400 episodes per language (380 for Chinese), distributed across the 14 games as
shown in Table~\ref{tab:episodes}. The counts follow the instance counts of the underlying
clembench release rather than being balanced by us, so games with more pre-compiled instances
contribute more episodes. The same instances are used for every model, so this affects the weight
of a game in the aggregate but not comparisons across models. Comparisons across languages are
affected in one case: dropping Wordle for Chinese removes a game on which every model scores
poorly, which raises Chinese scores by two to six points relative to the other 29 languages.

\begin{table}[h]
\centering
\small
\setlength{\tabcolsep}{5pt}
\begin{tabular}{@{}lr@{\hspace{14pt}}lr@{}}
\toprule
\textbf{Game} & \textbf{Ep.} & \textbf{Game} & \textbf{Ep.} \\
\midrule
TextMapWorld          & 50 & TextMapWorld (Room) & 30 \\
Clean Up              & 45 & Private \& Shared   & 25 \\
Image Game            & 40 & Match-It            & 20 \\
Codenames             & 35 & Taboo               & 20 \\
GuessWhat             & 30 & Wordle              & 20 \\
Reference Game        & 30 & Hot Air Balloon     & 15 \\
TextMapWorld (Graph)  & 30 & Deal or No Deal     & 10 \\
\midrule
\multicolumn{4}{@{}l}{\textbf{Total: 400} (380 for Chinese, where Wordle is not played)} \\
\bottomrule
\end{tabular}
\caption{Episodes per game in a single model--language run.}
\label{tab:episodes}
\end{table}

\section{Use of AI Assistants}
\label{app:ai-assistance}

AI coding assistants, primarily Codex (GPT-5.2) and Claude Sonnet 4, were used to write some of the code that produces the figures and tables in this paper. All experimental design, data collection and interpretation of results are the authors' own, and every reported number was verified against the underlying result files.

\section{Rank Correlation with External Benchmarks}
\label{sec:rank-correlation}

\begin{figure*}[t]
\centering
\begin{subfigure}[t]{0.46\textwidth}
    \centering
    \includegraphics[width=\linewidth]{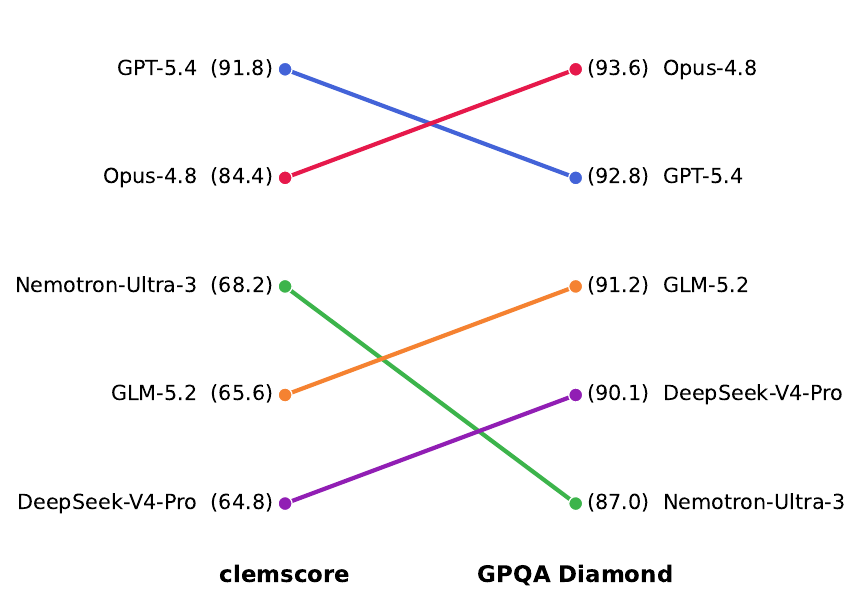}
    \caption{clemscore vs.\ GPQA Diamond ($\tau = 0.40$)}
\end{subfigure}\hfill
\begin{subfigure}[t]{0.46\textwidth}
    \centering
    \includegraphics[width=\linewidth]{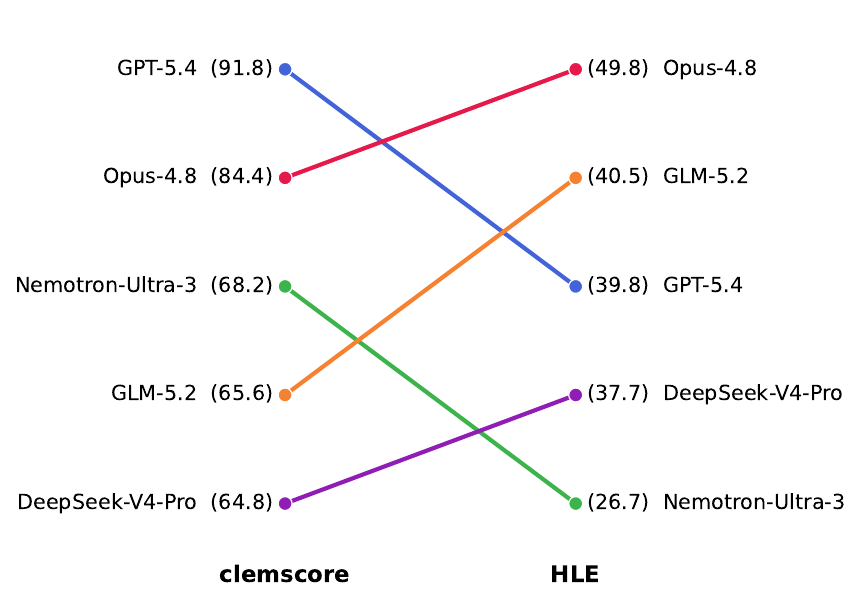}
    \caption{clemscore vs.\ HLE ($\tau = 0.20$)}
\end{subfigure}

\vspace{0.2em}
\begin{subfigure}[t]{0.46\textwidth}
    \centering
    \includegraphics[width=\linewidth]{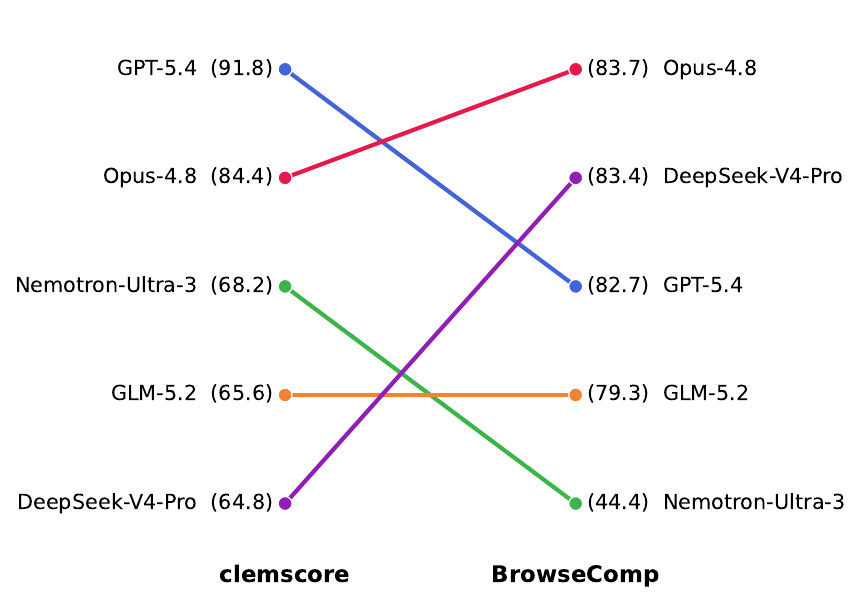}
    \caption{clemscore vs.\ BrowseComp ($\tau = 0.00$)}
\end{subfigure}\hfill
\begin{subfigure}[t]{0.46\textwidth}
    \centering
    \includegraphics[width=\linewidth]{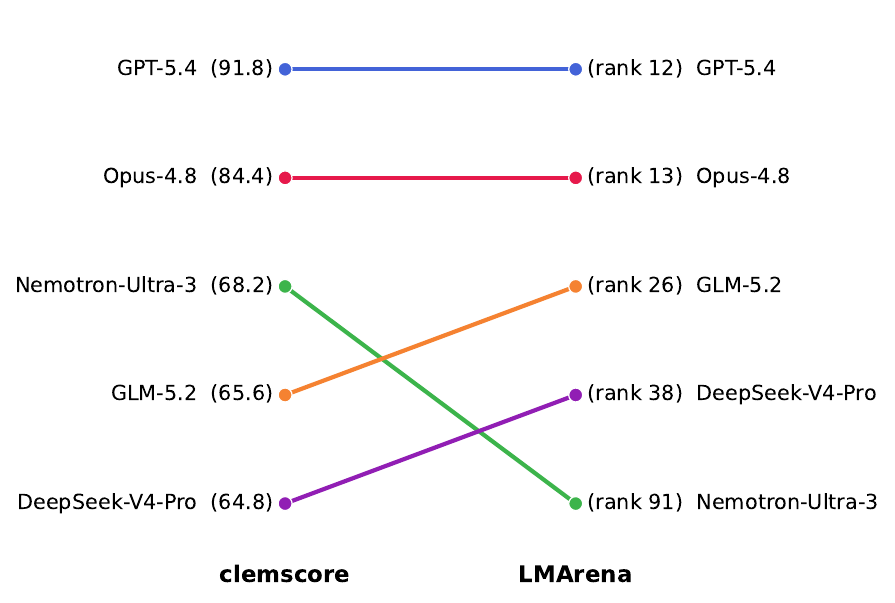}
    \caption{clemscore vs.\ LMArena ($\tau = 0.60$)}
\end{subfigure}
\caption{Ranking differences between clemscore (English, left column of each panel) and external benchmarks (right column), with Kendall's~$\tau$ per comparison.}
\label{fig:rank-correlation}
\end{figure*}

Figure~\ref{fig:rank-correlation} shows the ranking differences discussed in Section~\ref{sec:results} as bump charts, with agreement quantified by Kendall's~$\tau$: $0.60$ against LMArena, $0.40$ against GPQA Diamond, $0.20$ against HLE and $0.00$ against BrowseComp. Benchmark scores are taken from Table~\ref{tab:frontier-benchmarks}. With $n = 5$ these values are descriptive rather than significant.

Table~\ref{tab:frontier-benchmarks} reports benchmark scores for eight models across four capability clusters: knowledge and reasoning, mathematics, coding, and agentic task completion.
The remaining models in our study do not report results on these benchmarks and are omitted.
Scores were collected from each model's official release page or Hugging Face model card; cells marked ``---'' indicate that the model's authors did not report a result for that benchmark.
Note that MMLU-Pro (text knowledge) and MMMU-Pro (vision understanding) are distinct benchmarks, as are MCPMark and MCP Atlas.
Full model identifiers and source URLs for all nine models are listed in Table~\ref{tab:model-sources}.

\begin{table*}[t]
\centering
\setlength{\tabcolsep}{5.0pt}
\renewcommand{\arraystretch}{1.1}
\scriptsize
\begin{tabular}{@{}l *{7}{r} r@{}}
\toprule
\textbf{Benchmark}
  & \rotatebox{90}{\texttt{Qwen3.6}}
  & \rotatebox{90}{\texttt{Nemotron}}
  & \rotatebox{90}{\texttt{DS-V4-Pro}}
  & \rotatebox{90}{\texttt{GPT-5.4}}
  & \rotatebox{90}{\texttt{Opus-4.8}}
  & \rotatebox{90}{\texttt{GLM-5.2}}
  & \rotatebox{90}{\texttt{Gemma-4-26B}}
  & \rotatebox{90}{\texttt{Mistral-L3}} \\
\midrule
\multicolumn{9}{@{}l}{\rule{0pt}{8pt}\textit{Knowledge \& Reasoning}} \\[-3pt]
    \quad GPQA$^\diamond$        & 86.0  & 87.0  & 90.1  & 92.8  & 93.6           & 91.2           & 82.3  & 43.9 \\
    \quad HLE                    & 21.4  & 26.7  & 37.7  & 39.8  & 49.8           & 40.5           & 8.7  & {\color{gray}---} \\
    \quad HLE + tools            & {\color{gray}---} & 37.4  & 48.2  & 52.1  & 57.9           & 54.7           & 17.2  & {\color{gray}---} \\
    \quad MMLU-Pro               & 85.2  & {\color{gray}---} & 87.5  & {\color{gray}---} & {\color{gray}---} & {\color{gray}---} & 82.6  & {\color{gray}---} \\
    \quad MMMU-Pro               & 75.3  & 86.8  & {\color{gray}---} & 81.2  & {\color{gray}---} & {\color{gray}---} & 73.8  & {\color{gray}---} \\
    \quad MMLU                   & {\color{gray}---} & {\color{gray}---} & {\color{gray}---} & {\color{gray}---} & {\color{gray}---} & {\color{gray}---} & {\color{gray}---} & 85.5 \\
    \quad MMLU-Redux             & 93.3  & {\color{gray}---} & {\color{gray}---} & {\color{gray}---} & {\color{gray}---} & {\color{gray}---} & {\color{gray}---} & {\color{gray}---} \\
    \quad SuperGPQA              & 64.7  & {\color{gray}---} & {\color{gray}---} & {\color{gray}---} & {\color{gray}---} & {\color{gray}---} & {\color{gray}---} & {\color{gray}---} \\
    \quad Chinese-SimpleQA       & {\color{gray}---} & {\color{gray}---} & 84.4  & {\color{gray}---} & {\color{gray}---} & 75.0           & {\color{gray}---} & {\color{gray}---} \\
    \quad SimpleQA               & {\color{gray}---} & {\color{gray}---} & {\color{gray}---} & {\color{gray}---} & {\color{gray}---} & {\color{gray}---} & {\color{gray}---} & 23.8 \\
    \quad SimpleQA-Verified      & {\color{gray}---} & {\color{gray}---} & 57.9  & {\color{gray}---} & {\color{gray}---} & 38.1           & {\color{gray}---} & {\color{gray}---} \\
    \quad BBH                    & {\color{gray}---} & {\color{gray}---} & 87.5  & {\color{gray}---} & {\color{gray}---} & {\color{gray}---} & {\color{gray}---} & {\color{gray}---} \\
\multicolumn{9}{@{}l}{\rule{0pt}{8pt}\textit{Mathematics}} \\[-3pt]
    \quad AIME 2026              & 92.7  & {\color{gray}---} & {\color{gray}---} & {\color{gray}---} & 95.7           & 99.2           & 88.3  & {\color{gray}---} \\
    \quad HMMT Feb'26            & 83.6  & {\color{gray}---} & 95.2  & {\color{gray}---} & 96.7           & 92.5           & {\color{gray}---} & {\color{gray}---} \\
    \quad IMO-AB                 & 78.9  & 88.6  & 89.8  & {\color{gray}---} & 83.5           & 91.0           & {\color{gray}---} & {\color{gray}---} \\
    \quad IMO-AB + tools         & {\color{gray}---} & 92.3  & {\color{gray}---} & {\color{gray}---} & {\color{gray}---} & {\color{gray}---} & {\color{gray}---} & {\color{gray}---} \\
    \quad ARC-AGI-1              & {\color{gray}---} & {\color{gray}---} & {\color{gray}---} & 93.7  & {\color{gray}---} & {\color{gray}---} & {\color{gray}---} & {\color{gray}---} \\
    \quad ARC-AGI-2              & {\color{gray}---} & {\color{gray}---} & {\color{gray}---} & 73.3  & {\color{gray}---} & {\color{gray}---} & {\color{gray}---} & {\color{gray}---} \\
    \quad FrontierMath T1--3     & {\color{gray}---} & {\color{gray}---} & {\color{gray}---} & 47.6  & 43.8           & {\color{gray}---} & {\color{gray}---} & {\color{gray}---} \\
\multicolumn{9}{@{}l}{\rule{0pt}{8pt}\textit{Coding}} \\[-3pt]
    \quad SWE-Verified           & 73.4  & 71.9  & 80.6  & {\color{gray}---} & {\color{gray}---} & {\color{gray}---} & {\color{gray}---} & {\color{gray}---} \\
    \quad SWE-Pro                & 49.5  & {\color{gray}---} & 55.4  & 57.7  & 69.2           & 62.1           & {\color{gray}---} & {\color{gray}---} \\
    \quad SWE-Multi              & 67.2  & 67.7  & 76.2  & {\color{gray}---} & {\color{gray}---} & {\color{gray}---} & {\color{gray}---} & {\color{gray}---} \\
    \quad LiveCode v6            & 80.4  & 89.0  & {\color{gray}---} & {\color{gray}---} & 88.8           & {\color{gray}---} & 77.1  & {\color{gray}---} \\
    \quad LiveCodeBench          & {\color{gray}---} & {\color{gray}---} & 93.5  & {\color{gray}---} & {\color{gray}---} & {\color{gray}---} & {\color{gray}---} & 34.4 \\
    \quad Codeforces             & {\color{gray}---} & {\color{gray}---} & 3206$^\ddagger$ & 3168$^\ddagger$ & {\color{gray}---} & {\color{gray}---} & {\color{gray}---} & {\color{gray}---} \\
\multicolumn{9}{@{}l}{\rule{0pt}{8pt}\textit{Agentic}} \\[-3pt]
    \quad Terminal-B 2.0         & 51.5  & {\color{gray}---} & 67.9  & {\color{gray}---} & {\color{gray}---} & {\color{gray}---} & {\color{gray}---} & {\color{gray}---} \\
    \quad Terminal-B 2.1         & {\color{gray}---} & 56.4  & {\color{gray}---} & 75.1  & 85.0           & 81.0           & {\color{gray}---} & {\color{gray}---} \\
    \quad BrowseComp             & {\color{gray}---} & 44.4  & 83.4  & 82.7  & 83.7           & 79.3           & {\color{gray}---} & {\color{gray}---} \\
    \quad Toolathlon             & 26.9  & {\color{gray}---} & 51.8  & 54.6  & 59.9           & 48.2           & {\color{gray}---} & {\color{gray}---} \\
    \quad MCP Atlas              & 62.8  & {\color{gray}---} & 73.6  & 67.2  & 77.8           & 76.8           & {\color{gray}---} & {\color{gray}---} \\
    \quad MCPMark                & 37.0  & {\color{gray}---} & {\color{gray}---} & {\color{gray}---} & {\color{gray}---} & {\color{gray}---} & {\color{gray}---} & {\color{gray}---} \\
    \quad OSWorld                & {\color{gray}---} & {\color{gray}---} & {\color{gray}---} & 75.0  & {\color{gray}---} & {\color{gray}---} & {\color{gray}---} & {\color{gray}---} \\
    \quad GDPVal                 & {\color{gray}---} & 46.7  & {\color{gray}---} & 83.0  & {\color{gray}---} & {\color{gray}---} & {\color{gray}---} & {\color{gray}---} \\
\bottomrule
\end{tabular}
\caption{Benchmark scores for the eight models in this study with published results (scores in \%; Codeforces reported as Elo rating).
  \textbf{Qwen3.6}~=~Qwen3.6-35B-A3B;
  \textbf{Nemotron}~=~Nemotron-3-Ultra-550B-A55B;
  \textbf{DS-V4-Pro}~=~DeepSeek-V4-Pro;
  \textbf{Opus-4.8}~=~Claude-Opus-4.8;
  \textbf{Gemma-4-26B}~=~Gemma-4-26b-a4b-it;
  \textbf{Mistral-L3}~=~Mistral-Large-3-675B-Instruct-2512.
  GPQA$^\diamond$~=~GPQA Diamond; HLE~=~Humanity's Last Exam;
  MMLU-Pro~=~Massive Multitask Language Understanding Professional (text);
  MMMU-Pro~=~Massive Multitask Multimodal Understanding Pro (vision);
  BBH~=~BIG-Bench Hard;
  IMO-AB~=~IMOAnswerBench; SWE-Multi~=~SWE-Bench Multilingual;
  LiveCode~v6~=~LiveCodeBench~v6; LiveCodeBench~=~LiveCodeBench (no CoT).
  $^\ddagger$~Elo rating rather than a percentage.
  Evaluation results for Apertus-1.5 are not available yet as of July 2026. All figures are vendor-self-reported under each vendor's own evaluation protocol, so they indicate approximate standing rather than strictly comparable measurements.}
\label{tab:frontier-benchmarks}
\end{table*}

\begin{table}[t]
\centering
\scriptsize
\setlength{\tabcolsep}{3pt}
\begin{tabular}{@{}l >{\raggedright\arraybackslash}p{3.35cm} r@{}}
\toprule
\textbf{Model} & \textbf{Source} & \textbf{\$ in\,/\,out} \\
\midrule
\texttt{GPT-5.4}         & \url{openai.com/index/gpt-5/}                & 2.50 / 15.00 \\
\texttt{Claude-Opus-4.8} & \url{anthropic.com/claude/opus}              & 5.00 / 25.00 \\
\texttt{GLM-5.2}         & \url{hf.co/zai-org/GLM-5.2}                  & 1.00 / 4.00 \\
\texttt{Nemotron-3-Ultra} & \url{hf.co/nvidia/NVIDIA-Nemotron-3-Ultra-550B-A55B-BF16} & 0.50 / 2.20 \\
\texttt{DeepSeek-V4-Pro} & \url{hf.co/deepseek-ai/DeepSeek-V4-Pro}      & 0.44 / 0.87 \\
\texttt{Mistral-Large-3} & \url{hf.co/mistralai/Mistral-Large-3-675B-Instruct-2512} & 0.50 / 1.50 \\
\texttt{Qwen3.6-35B-A3B} & \url{hf.co/Qwen/Qwen3.6-35B-A3B}             & 0.32 / 1.28 \\
\texttt{Gemma-4-26B}     & \url{hf.co/google/gemma-4-26B-A4B-it}        & 0.07 / 0.35 \\
\texttt{Apertus-1.5-8B}  & \url{hf.co/swiss-ai/Apertus-v1.5-8B}         & 0.07 / 0.15 \\
\bottomrule
\end{tabular}
\caption{Model identifiers, sources and API pricing (USD per million input\,/\,output tokens); \texttt{hf.co} abbreviates \texttt{huggingface.co}. Benchmark scores were collected from each model's official release page or Hugging Face model card, and pricing from OpenRouter in July 2026. Apertus-1.5 is not listed on OpenRouter, so we used the price of Qwen-3-8B, the closest model by size.}
\label{tab:model-sources}
\end{table}

\section{Additional Results}
\label{app:results}
This appendix reports the per-language detail behind Section~\ref{sec:results}: estimated API cost (Table~\ref{tab:cost-per-language}), the two clemscore components \%Played and Quality separately (Tables~\ref{tab:played-per-language} and~\ref{tab:quality-per-language}), and clemscore broken down by game and model within each capability cluster (Tables~\ref{tab:cgmT-lexical}--\ref{tab:cgmT-spatial}).

Costs are computed from the token usage recorded in the interaction transcripts and each provider's per-token pricing, over the full benchmark run, which is the quantity a deployment actually pays. Three of the nine models were run locally (Section~\ref{sec:models}), so their costs are OpenRouter list prices rather than amounts billed. The aggregate also does not separate models that complete episodes from models that abort early: a model that terminates a game prematurely consumes fewer tokens and therefore appears cheaper, so cost should be read alongside \%Played rather than on its own.

\begin{table}[t!]
\centering
\scriptsize
\setlength{\tabcolsep}{2.5pt}
\renewcommand{\arraystretch}{1.05}
\begin{tabular}{@{}l *{9}{r} *{1}{r}@{}}
\toprule
\textbf{Lang}
  & \rotatebox{90}{\texttt{GPT-5.4}}
  & \rotatebox{90}{\texttt{Opus-4.8}}
  & \rotatebox{90}{\texttt{GLM-5.2}}
  & \rotatebox{90}{\texttt{DS-V4}}
  & \rotatebox{90}{\texttt{Nemotron}}
  & \rotatebox{90}{\texttt{Gemma-4}}
  & \rotatebox{90}{\texttt{Qwen3.6}}
  & \rotatebox{90}{\texttt{Mistral-L3}}
  & \rotatebox{90}{\texttt{Apertus}}
  & \rotatebox{90}{\textbf{Avg}} \\
\midrule
ar & \cellcolor{orange!52}31.7 & \cellcolor{orange!52}38.1 & \cellcolor{orange!36}9.34 & \cellcolor{orange!28}3.16 & \cellcolor{orange!28}3.19 & \cellcolor{orange!5}0.47 & \cellcolor{orange!20}1.53 & \cellcolor{orange!20}1.77 & \cellcolor{orange!5}0.41 & \cellcolor{orange!36}9.97 \\
\rowcolor{gray!12} bg & \cellcolor{orange!52}32.3 & \cellcolor{orange!52}36.3 & \cellcolor{orange!36}7.76 & \cellcolor{orange!28}3.26 & \cellcolor{orange!28}3.37 & \cellcolor{orange!13}0.70 & \cellcolor{orange!28}2.59 & \cellcolor{orange!20}1.76 & \cellcolor{orange!28}2.48 & \cellcolor{orange!36}10.1 \\
cs & \cellcolor{orange!52}31.0 & \cellcolor{orange!52}38.7 & \cellcolor{orange!36}7.90 & \cellcolor{orange!28}3.49 & \cellcolor{orange!28}4.26 & \cellcolor{orange!13}0.69 & \cellcolor{orange!20}2.15 & \cellcolor{orange!20}2.18 & \cellcolor{orange!28}3.27 & \cellcolor{orange!44}10.4 \\
\rowcolor{gray!12} da & \cellcolor{orange!52}28.3 & \cellcolor{orange!52}36.5 & \cellcolor{orange!36}8.61 & \cellcolor{orange!28}3.80 & \cellcolor{orange!28}4.53 & \cellcolor{orange!13}0.62 & \cellcolor{orange!28}2.79 & \cellcolor{orange!20}2.27 & \cellcolor{orange!5}0.59 & \cellcolor{orange!36}9.78 \\
de & \cellcolor{orange!52}29.4 & \cellcolor{orange!52}41.9 & \cellcolor{orange!36}7.09 & \cellcolor{orange!28}3.67 & \cellcolor{orange!28}3.97 & \cellcolor{orange!5}0.54 & \cellcolor{orange!28}2.50 & \cellcolor{orange!20}2.44 & \cellcolor{orange!20}2.31 & \cellcolor{orange!44}10.4 \\
\rowcolor{gray!12} el & \cellcolor{orange!52}34.7 & \cellcolor{orange!60}50.4 & \cellcolor{orange!36}9.68 & \cellcolor{orange!36}5.49 & \cellcolor{orange!36}6.14 & \cellcolor{orange!13}0.72 & \cellcolor{orange!28}3.16 & \cellcolor{orange!20}1.81 & \cellcolor{orange!20}1.55 & \cellcolor{orange!44}12.6 \\
en & \cellcolor{orange!52}24.7 & \cellcolor{orange!52}30.6 & \cellcolor{orange!36}5.22 & \cellcolor{orange!28}2.67 & \cellcolor{orange!28}3.31 & \cellcolor{orange!5}0.52 & \cellcolor{orange!20}1.93 & \cellcolor{orange!20}2.00 & \cellcolor{orange!5}0.41 & \cellcolor{orange!36}7.93 \\
\rowcolor{gray!12} es & \cellcolor{orange!52}33.2 & \cellcolor{orange!52}38.7 & \cellcolor{orange!28}4.35 & \cellcolor{orange!20}2.37 & \cellcolor{orange!20}2.45 & \cellcolor{orange!5}0.30 & \cellcolor{orange!20}1.66 & \cellcolor{orange!20}1.50 & \cellcolor{orange!5}0.36 & \cellcolor{orange!36}9.43 \\
et & \cellcolor{orange!52}31.1 & \cellcolor{orange!52}40.5 & \cellcolor{orange!36}9.11 & \cellcolor{orange!28}4.32 & \cellcolor{orange!28}4.04 & \cellcolor{orange!13}0.63 & \cellcolor{orange!28}2.51 & \cellcolor{orange!20}1.55 & \cellcolor{orange!20}1.68 & \cellcolor{orange!44}10.6 \\
\rowcolor{gray!12} fi & \cellcolor{orange!52}30.2 & \cellcolor{orange!52}41.4 & \cellcolor{orange!36}8.43 & \cellcolor{orange!28}3.93 & \cellcolor{orange!28}4.50 & \cellcolor{orange!5}0.48 & \cellcolor{orange!20}2.38 & \cellcolor{orange!28}2.70 & \cellcolor{orange!28}4.10 & \cellcolor{orange!44}10.9 \\
fr & \cellcolor{orange!52}29.7 & \cellcolor{orange!52}37.3 & \cellcolor{orange!36}7.09 & \cellcolor{orange!28}3.50 & \cellcolor{orange!28}4.04 & \cellcolor{orange!5}0.58 & \cellcolor{orange!20}1.96 & \cellcolor{orange!20}1.74 & \cellcolor{orange!13}0.93 & \cellcolor{orange!36}9.66 \\
\rowcolor{gray!12} ga & \cellcolor{orange!60}53.8 & \cellcolor{orange!60}82.5 & \cellcolor{orange!44}17.4 & \cellcolor{orange!36}5.10 & \cellcolor{orange!36}6.80 & \cellcolor{orange!13}1.00 & \cellcolor{orange!36}5.16 & \cellcolor{orange!20}1.95 & \cellcolor{orange!36}6.15 & \cellcolor{orange!44}20.0 \\
hr & \cellcolor{orange!52}31.4 & \cellcolor{orange!52}37.5 & \cellcolor{orange!36}8.39 & \cellcolor{orange!28}3.78 & \cellcolor{orange!28}4.23 & \cellcolor{orange!13}0.64 & \cellcolor{orange!20}1.71 & \cellcolor{orange!20}2.10 & \cellcolor{orange!13}0.95 & \cellcolor{orange!36}10.1 \\
\rowcolor{gray!12} hu & \cellcolor{orange!52}32.7 & \cellcolor{orange!52}42.0 & \cellcolor{orange!44}10.6 & \cellcolor{orange!28}4.07 & \cellcolor{orange!28}4.45 & \cellcolor{orange!13}0.64 & \cellcolor{orange!28}2.87 & \cellcolor{orange!20}2.10 & \cellcolor{orange!5}0.50 & \cellcolor{orange!44}11.1 \\
it & \cellcolor{orange!52}28.4 & \cellcolor{orange!52}36.7 & \cellcolor{orange!36}8.12 & \cellcolor{orange!28}3.11 & \cellcolor{orange!28}3.73 & \cellcolor{orange!13}0.61 & \cellcolor{orange!20}2.45 & \cellcolor{orange!20}1.72 & \cellcolor{orange!20}2.27 & \cellcolor{orange!36}9.67 \\
\rowcolor{gray!12} lt & \cellcolor{orange!60}53.6 & \cellcolor{orange!60}85.3 & \cellcolor{orange!44}14.2 & \cellcolor{orange!28}3.59 & \cellcolor{orange!36}5.44 & \cellcolor{orange!5}0.58 & \cellcolor{orange!28}4.52 & \cellcolor{orange!20}1.75 & \cellcolor{orange!13}0.74 & \cellcolor{orange!44}18.9 \\
lv & \cellcolor{orange!52}32.1 & \cellcolor{orange!52}40.1 & \cellcolor{orange!36}10.2 & \cellcolor{orange!28}4.40 & \cellcolor{orange!36}5.17 & \cellcolor{orange!13}0.79 & \cellcolor{orange!20}2.35 & \cellcolor{orange!20}1.94 & \cellcolor{orange!36}9.38 & \cellcolor{orange!44}11.8 \\
\rowcolor{gray!12} mt & \cellcolor{orange!52}33.1 & \cellcolor{orange!60}44.2 & \cellcolor{orange!44}11.8 & \cellcolor{orange!36}5.03 & \cellcolor{orange!36}8.32 & \cellcolor{orange!13}0.67 & \cellcolor{orange!28}4.98 & \cellcolor{orange!20}1.85 & \cellcolor{orange!5}0.54 & \cellcolor{orange!44}12.3 \\
nl & \cellcolor{orange!52}28.4 & \cellcolor{orange!52}36.8 & \cellcolor{orange!36}6.52 & \cellcolor{orange!28}3.71 & \cellcolor{orange!28}3.89 & \cellcolor{orange!5}0.57 & \cellcolor{orange!28}3.09 & \cellcolor{orange!28}2.91 & \cellcolor{orange!28}4.74 & \cellcolor{orange!36}10.1 \\
\rowcolor{gray!12} pl & \cellcolor{orange!52}31.3 & \cellcolor{orange!52}40.5 & \cellcolor{orange!36}9.00 & \cellcolor{orange!28}3.75 & \cellcolor{orange!28}4.54 & \cellcolor{orange!5}0.58 & \cellcolor{orange!28}2.75 & \cellcolor{orange!28}2.96 & \cellcolor{orange!13}0.71 & \cellcolor{orange!44}10.7 \\
pt & \cellcolor{orange!52}35.7 & \cellcolor{orange!52}38.6 & \cellcolor{orange!36}5.41 & \cellcolor{orange!20}2.22 & \cellcolor{orange!20}2.42 & \cellcolor{orange!5}0.30 & \cellcolor{orange!20}1.61 & \cellcolor{orange!20}1.23 & \cellcolor{orange!5}0.31 & \cellcolor{orange!36}9.76 \\
\rowcolor{gray!12} ro & \cellcolor{orange!52}34.1 & \cellcolor{orange!52}38.4 & \cellcolor{orange!36}8.39 & \cellcolor{orange!28}4.30 & \cellcolor{orange!28}4.36 & \cellcolor{orange!13}0.67 & \cellcolor{orange!28}3.01 & \cellcolor{orange!28}2.52 & \cellcolor{orange!13}0.64 & \cellcolor{orange!44}10.7 \\
ru & \cellcolor{orange!52}31.7 & \cellcolor{orange!52}36.8 & \cellcolor{orange!36}7.05 & \cellcolor{orange!28}4.01 & \cellcolor{orange!36}5.24 & \cellcolor{orange!5}0.48 & \cellcolor{orange!20}2.33 & \cellcolor{orange!20}2.47 & \cellcolor{orange!36}6.31 & \cellcolor{orange!44}10.7 \\
\rowcolor{gray!12} sk & \cellcolor{orange!52}31.8 & \cellcolor{orange!52}39.8 & \cellcolor{orange!36}7.52 & \cellcolor{orange!28}3.85 & \cellcolor{orange!36}5.10 & \cellcolor{orange!13}0.61 & \cellcolor{orange!28}3.09 & \cellcolor{orange!28}2.51 & \cellcolor{orange!5}0.38 & \cellcolor{orange!44}10.5 \\
sl & \cellcolor{orange!52}30.5 & \cellcolor{orange!52}38.0 & \cellcolor{orange!36}8.68 & \cellcolor{orange!28}3.38 & \cellcolor{orange!28}4.21 & \cellcolor{orange!13}0.66 & \cellcolor{orange!28}2.50 & \cellcolor{orange!20}1.81 & \cellcolor{orange!20}1.23 & \cellcolor{orange!36}10.1 \\
\rowcolor{gray!12} sr & \cellcolor{orange!52}34.2 & \cellcolor{orange!52}36.6 & \cellcolor{orange!36}8.72 & \cellcolor{orange!28}3.70 & \cellcolor{orange!28}4.28 & \cellcolor{orange!5}0.57 & \cellcolor{orange!20}2.15 & \cellcolor{orange!20}2.09 & \cellcolor{orange!13}0.60 & \cellcolor{orange!44}10.3 \\
sv & \cellcolor{orange!52}27.7 & \cellcolor{orange!52}35.9 & \cellcolor{orange!36}7.19 & \cellcolor{orange!28}3.49 & \cellcolor{orange!28}4.12 & \cellcolor{orange!13}0.65 & \cellcolor{orange!28}2.85 & \cellcolor{orange!28}2.56 & \cellcolor{orange!20}2.16 & \cellcolor{orange!36}9.62 \\
\rowcolor{gray!12} tr & \cellcolor{orange!52}35.0 & \cellcolor{orange!60}42.4 & \cellcolor{orange!44}12.6 & \cellcolor{orange!28}4.46 & \cellcolor{orange!28}4.81 & \cellcolor{orange!13}0.68 & \cellcolor{orange!28}3.60 & \cellcolor{orange!13}1.08 & \cellcolor{orange!28}3.57 & \cellcolor{orange!44}12.0 \\
uk & \cellcolor{orange!52}32.0 & \cellcolor{orange!52}38.3 & \cellcolor{orange!36}7.32 & \cellcolor{orange!28}4.39 & \cellcolor{orange!28}4.87 & \cellcolor{orange!5}0.53 & \cellcolor{orange!28}3.10 & \cellcolor{orange!28}2.72 & \cellcolor{orange!20}1.37 & \cellcolor{orange!44}10.5 \\
\rowcolor{gray!12} zh & \cellcolor{orange!52}26.1 & \cellcolor{orange!52}26.2 & \cellcolor{orange!20}1.78 & \cellcolor{orange!5}0.59 & \cellcolor{orange!20}1.29 & \cellcolor{orange!5}0.48 & \cellcolor{orange!20}2.33 & \cellcolor{orange!28}2.92 & \cellcolor{orange!20}1.67 & \cellcolor{orange!36}7.04 \\
\midrule
Total & \cellcolor{orange!60}980.3 & \cellcolor{orange!60}1246.9 & \cellcolor{orange!60}255.4 & \cellcolor{orange!60}110.5 & \cellcolor{orange!60}131.1 & \cellcolor{orange!44}18.0 & \cellcolor{orange!60}81.6 & \cellcolor{orange!60}62.9 & \cellcolor{orange!60}62.3 & \cellcolor{orange!60}2949.0 \\
Avg & \cellcolor{orange!52}32.7 & \cellcolor{orange!52}41.6 & \cellcolor{orange!36}8.51 & \cellcolor{orange!28}3.68 & \cellcolor{orange!28}4.37 & \cellcolor{orange!5}0.60 & \cellcolor{orange!28}2.72 & \cellcolor{orange!20}2.10 & \cellcolor{orange!20}2.08 & \cellcolor{orange!44}10.9 \\
STD & \cellcolor{orange!36}6.15 & \cellcolor{orange!44}12.0 & \cellcolor{orange!28}2.87 & \cellcolor{orange!13}0.92 & \cellcolor{orange!20}1.29 & \cellcolor{orange!60}0.13 & \cellcolor{orange!13}0.88 & \cellcolor{orange!5}0.48 & \cellcolor{orange!20}2.14 & \cellcolor{orange!44}15.1 \\
\bottomrule
\end{tabular}%
\caption{API cost in \$USD to run the full benchmark per language (rows) and model (columns); columns ordered by mean clemscore. The Total row is the summed cost across all languages. Cell shading scales with log-cost; the cheapest model per language is in \textbf{bold}.}
\label{tab:cost-per-language}
\end{table}

% Both tables in a single float, side by side, so they always stay
% horizontally aligned next to each other when compiled.
\begin{table*}[t!]
\centering
\begin{minipage}[t]{0.49\textwidth}
\centering
\scriptsize
\setlength{\tabcolsep}{2.5pt}
\renewcommand{\arraystretch}{1.05}
\resizebox{\linewidth}{!}{%
\begin{tabular}{@{}l *{9}{r} *{1}{r}@{}}
\toprule
\textbf{Lang}
  & \rotatebox{90}{\texttt{GPT-5.4}}
  & \rotatebox{90}{\texttt{Opus-4.8}}
  & \rotatebox{90}{\texttt{GLM-5.2}}
  & \rotatebox{90}{\texttt{DS-V4}}
  & \rotatebox{90}{\texttt{Nemotron}}
  & \rotatebox{90}{\texttt{Gemma-4}}
  & \rotatebox{90}{\texttt{Qwen3.6}}
  & \rotatebox{90}{\texttt{Mistral-L3}}
  & \rotatebox{90}{\texttt{Apertus}}
  & \rotatebox{90}{\textbf{Avg}} \\
\midrule
ar & \cellcolor{violet!44}89.0 & \cellcolor{violet!28}83.5 & \cellcolor{violet!52}89.6 & \cellcolor{violet!20}78.7 & \cellcolor{violet!13}72.8 & \cellcolor{violet!13}71.9 & \cellcolor{violet!13}70.9 & \cellcolor{violet!13}67.8 & \cellcolor{violet!5}40.4 & \cellcolor{violet!13}73.8 \\
\rowcolor{gray!12} bg & \cellcolor{violet!60}99.2 & \cellcolor{violet!52}93.2 & \cellcolor{violet!52}89.9 & \cellcolor{violet!28}81.9 & \cellcolor{violet!20}79.0 & \cellcolor{violet!36}85.1 & \cellcolor{violet!44}87.5 & \cellcolor{violet!13}69.6 & \cellcolor{violet!5}48.7 & \cellcolor{violet!28}81.6 \\
cs & \cellcolor{violet!60}98.8 & \cellcolor{violet!52}91.4 & \cellcolor{violet!20}80.5 & \cellcolor{violet!36}86.5 & \cellcolor{violet!28}82.2 & \cellcolor{violet!20}80.3 & \cellcolor{violet!52}89.4 & \cellcolor{violet!13}68.9 & \cellcolor{violet!5}44.0 & \cellcolor{violet!20}80.2 \\
\rowcolor{gray!12} da & \cellcolor{violet!60}97.4 & \cellcolor{violet!52}93.3 & \cellcolor{violet!52}90.6 & \cellcolor{violet!52}90.9 & \cellcolor{violet!44}88.2 & \cellcolor{violet!44}88.8 & \cellcolor{violet!36}85.4 & \cellcolor{violet!13}73.7 & \cellcolor{violet!5}41.5 & \cellcolor{violet!28}83.3 \\
de & \cellcolor{violet!60}97.9 & \cellcolor{violet!60}94.6 & \cellcolor{violet!52}92.4 & \cellcolor{violet!60}95.1 & \cellcolor{violet!36}86.6 & \cellcolor{violet!44}87.1 & \cellcolor{violet!20}79.8 & \cellcolor{violet!20}78.4 & \cellcolor{violet!5}50.4 & \cellcolor{violet!28}84.7 \\
\bottomrule
\end{tabular}%
}
\caption{Percentage of successfully played episodes (\% Played) per language (rows) and model (columns), averaged over the 14 games; columns ordered by mean clemscore. Cells are shaded by octile of the distribution (darker = higher); the best model per language is in \textbf{bold}.}
\label{tab:played-per-language}
\end{minipage}\hfill
\begin{minipage}[t]{0.49\textwidth}
\centering
\scriptsize
\setlength{\tabcolsep}{2.5pt}
\renewcommand{\arraystretch}{1.05}
\resizebox{\linewidth}{!}{%
\begin{tabular}{@{}l *{9}{r} *{1}{r}@{}}
\toprule
\textbf{Lang}
  & \rotatebox{90}{\texttt{GPT-5.4}}
  & \rotatebox{90}{\texttt{Opus-4.8}}
  & \rotatebox{90}{\texttt{GLM-5.2}}
  & \rotatebox{90}{\texttt{DS-V4}}
  & \rotatebox{90}{\texttt{Nemotron}}
  & \rotatebox{90}{\texttt{Gemma-4}}
  & \rotatebox{90}{\texttt{Qwen3.6}}
  & \rotatebox{90}{\texttt{Mistral-L3}}
  & \rotatebox{90}{\texttt{Apertus}}
  & \rotatebox{90}{\textbf{Avg}} \\
\midrule
ar & \cellcolor{teal!52}88.9 & \cellcolor{teal!52}82.8 & \cellcolor{teal!36}67.2 & \cellcolor{teal!28}66.2 & \cellcolor{teal!28}64.7 & \cellcolor{teal!36}69.4 & \cellcolor{teal!13}48.9 & \cellcolor{teal!20}58.0 & \cellcolor{teal!5}29.1 & \cellcolor{teal!28}63.9 \\
\rowcolor{gray!12} bg & \cellcolor{teal!60}93.2 & \cellcolor{teal!52}86.5 & \cellcolor{teal!52}76.7 & \cellcolor{teal!52}74.7 & \cellcolor{teal!28}66.2 & \cellcolor{teal!44}71.4 & \cellcolor{teal!20}60.0 & \cellcolor{teal!20}59.9 & \cellcolor{teal!5}36.1 & \cellcolor{teal!36}69.4 \\
cs & \cellcolor{teal!60}91.7 & \cellcolor{teal!60}89.4 & \cellcolor{teal!52}74.7 & \cellcolor{teal!36}67.8 & \cellcolor{teal!36}67.8 & \cellcolor{teal!28}65.7 & \cellcolor{teal!13}56.4 & \cellcolor{teal!20}60.7 & \cellcolor{teal!5}32.1 & \cellcolor{teal!36}67.4 \\
\rowcolor{gray!12} da & \cellcolor{teal!60}92.4 & \cellcolor{teal!52}89.0 & \cellcolor{teal!52}75.6 & \cellcolor{teal!44}71.5 & \cellcolor{teal!28}66.3 & \cellcolor{teal!36}66.6 & \cellcolor{teal!13}55.5 & \cellcolor{teal!20}60.8 & \cellcolor{teal!5}27.2 & \cellcolor{teal!36}67.2 \\
de & \cellcolor{teal!60}92.7 & \cellcolor{teal!60}90.5 & \cellcolor{teal!44}72.9 & \cellcolor{teal!44}73.7 & \cellcolor{teal!44}72.7 & \cellcolor{teal!36}66.5 & \cellcolor{teal!13}53.3 & \cellcolor{teal!28}63.0 & \cellcolor{teal!5}32.2 & \cellcolor{teal!36}68.6 \\
\bottomrule
\end{tabular}%
}
\caption{Quality Score of successfully played episodes per language (rows) and model (columns), averaged over the 14 games; columns ordered by mean clemscore. Cells are shaded by octile of the distribution (darker = higher); the best model per language is in \textbf{bold}.}
\label{tab:quality-per-language}
\end{minipage}
\end{table*}

\begin{table*}[t]
\centering
\scriptsize
\setlength{\tabcolsep}{2.2pt}
\renewcommand{\arraystretch}{1.05}
\resizebox{\textwidth}{!}{%
\begin{tabular}{@{}ll *{30}{c}@{}}
\toprule
\textbf{Game} & \textbf{Model} & \textbf{ar} & \textbf{bg} & \textbf{cs} & \textbf{da} & \textbf{de} & \textbf{el} & \textbf{en} & \textbf{es} & \textbf{et} & \textbf{fi} & \textbf{fr} & \textbf{ga} & \textbf{hr} & \textbf{hu} & \textbf{it} & \textbf{lt} & \textbf{lv} & \textbf{mt} & \textbf{nl} & \textbf{pl} & \textbf{pt} & \textbf{ro} & \textbf{ru} & \textbf{sk} & \textbf{sl} & \textbf{sr} & \textbf{sv} & \textbf{tr} & \textbf{uk} & \textbf{zh} \\
\midrule
\multirow{9}{*}{\texttt{taboo}} & \texttt{GPT} & \cellcolor{teal!40}58 & \cellcolor{teal!55}98 & \cellcolor{teal!55}95 & \cellcolor{teal!47}82 & \cellcolor{teal!47}79 & \cellcolor{teal!55}90 & \cellcolor{teal!55}95 & \cellcolor{teal!55}92 & \cellcolor{teal!47}80 & \cellcolor{teal!55}98 & \cellcolor{teal!55}98 & \cellcolor{teal!55}88 & \cellcolor{teal!55}95 & \cellcolor{teal!55}95 & \cellcolor{teal!55}98 & \cellcolor{teal!55}100 & \cellcolor{teal!55}97 & \cellcolor{teal!47}82 & \cellcolor{teal!55}90 & \cellcolor{teal!55}92 & \cellcolor{teal!55}85 & \cellcolor{teal!55}98 & \cellcolor{teal!55}92 & \cellcolor{teal!47}79 & \cellcolor{teal!55}90 & \cellcolor{teal!55}98 & \cellcolor{teal!55}90 & \cellcolor{teal!55}88 & \cellcolor{teal!55}98 & \cellcolor{teal!55}93 \\
\bottomrule
\end{tabular}%
}
\caption{clemscore for the \textbf{Lexical / world knowledge} capability cluster: one row per game and model, one column per language. Shading uses this cluster's own score distribution (equal-population octiles, darker = higher); ``-'' marks a missing run. Models: \texttt{GPT}~=~GPT-5.4; \texttt{Opus}~=~Claude Opus 4.8; \texttt{GLM}~=~GLM-5.2; \texttt{DSV4}~=~DeepSeek-V4-Pro; \texttt{Nemo}~=~Nemotron-3-Ultra; \texttt{Gem4}~=~Gemma-4-26B; \texttt{Qwen}~=~Qwen3.6-35B-A3B; \texttt{Mist}~=~Mistral Large 3; \texttt{Aper}~=~Apertus-v1.5-8B.}
\label{tab:cgmT-lexical}
\end{table*}

\begin{table*}[t]
\centering
\scriptsize
\setlength{\tabcolsep}{2.2pt}
\renewcommand{\arraystretch}{1.05}
\resizebox{\textwidth}{!}{%
\begin{tabular}{@{}ll *{30}{c}@{}}
\toprule
\textbf{Game} & \textbf{Model} & \textbf{ar} & \textbf{bg} & \textbf{cs} & \textbf{da} & \textbf{de} & \textbf{el} & \textbf{en} & \textbf{es} & \textbf{et} & \textbf{fi} & \textbf{fr} & \textbf{ga} & \textbf{hr} & \textbf{hu} & \textbf{it} & \textbf{lt} & \textbf{lv} & \textbf{mt} & \textbf{nl} & \textbf{pl} & \textbf{pt} & \textbf{ro} & \textbf{ru} & \textbf{sk} & \textbf{sl} & \textbf{sr} & \textbf{sv} & \textbf{tr} & \textbf{uk} & \textbf{zh} \\
\midrule
\multirow{9}{*}{\texttt{reference}} & \texttt{GPT} & \cellcolor{orange!55}100 & \cellcolor{orange!55}100 & \cellcolor{orange!55}100 & \cellcolor{orange!55}100 & \cellcolor{orange!55}100 & \cellcolor{orange!55}100 & \cellcolor{orange!55}100 & \cellcolor{orange!55}100 & \cellcolor{orange!55}100 & \cellcolor{orange!55}100 & \cellcolor{orange!55}100 & \cellcolor{orange!5}3 & \cellcolor{orange!55}100 & \cellcolor{orange!12}40 & \cellcolor{orange!55}100 & \cellcolor{orange!5}10 & \cellcolor{orange!55}100 & \cellcolor{orange!55}100 & \cellcolor{orange!55}100 & \cellcolor{orange!55}100 & \cellcolor{orange!55}100 & \cellcolor{orange!5}13 & \cellcolor{orange!55}100 & \cellcolor{orange!55}100 & \cellcolor{orange!12}30 & \cellcolor{orange!55}100 & \cellcolor{orange!55}100 & \cellcolor{orange!55}100 & \cellcolor{orange!55}100 & \cellcolor{orange!55}100 \\
 & \texttt{Opus} & \cellcolor{orange!47}97 & \cellcolor{orange!55}100 & \cellcolor{orange!47}97 & \cellcolor{orange!40}90 & \cellcolor{orange!47}97 & \cellcolor{orange!55}100 & \cellcolor{orange!55}100 & \cellcolor{orange!55}100 & \cellcolor{orange!47}97 & \cellcolor{orange!47}97 & \cellcolor{orange!55}100 & \cellcolor{orange!47}97 & \cellcolor{orange!47}96 & \cellcolor{orange!55}100 & \cellcolor{orange!55}100 & \cellcolor{orange!47}97 & \cellcolor{orange!55}100 & \cellcolor{orange!40}93 & \cellcolor{orange!47}97 & \cellcolor{orange!55}100 & \cellcolor{orange!55}100 & \cellcolor{orange!47}97 & \cellcolor{orange!55}100 & \cellcolor{orange!55}100 & \cellcolor{orange!47}93 & \cellcolor{orange!55}100 & \cellcolor{orange!26}80 & \cellcolor{orange!55}100 & \cellcolor{orange!47}97 & \cellcolor{orange!55}100 \\
 & \texttt{GLM} & \cellcolor{orange!33}83 & \cellcolor{orange!40}90 & \cellcolor{orange!26}77 & \cellcolor{orange!19}67 & \cellcolor{orange!19}67 & \cellcolor{orange!19}73 & \cellcolor{orange!40}93 & \cellcolor{orange!26}80 & \cellcolor{orange!26}80 & \cellcolor{orange!26}80 & \cellcolor{orange!40}87 & \cellcolor{orange!12}48 & \cellcolor{orange!26}80 & \cellcolor{orange!12}57 & \cellcolor{orange!26}80 & \cellcolor{orange!33}80 & \cellcolor{orange!26}80 & \cellcolor{orange!33}83 & \cellcolor{orange!33}86 & \cellcolor{orange!19}73 & \cellcolor{orange!26}77 & \cellcolor{orange!12}53 & \cellcolor{orange!26}77 & \cellcolor{orange!33}83 & \cellcolor{orange!12}60 & \cellcolor{orange!33}80 & \cellcolor{orange!33}83 & \cellcolor{orange!47}93 & \cellcolor{orange!47}93 & \cellcolor{orange!33}83 \\
 & \texttt{DSV4} & \cellcolor{orange!33}83 & \cellcolor{orange!19}73 & \cellcolor{orange!33}83 & \cellcolor{orange!19}73 & \cellcolor{orange!47}93 & \cellcolor{orange!12}47 & \cellcolor{orange!33}80 & \cellcolor{orange!26}80 & \cellcolor{orange!12}60 & \cellcolor{orange!12}53 & \cellcolor{orange!5}0 & \cellcolor{orange!12}33 & \cellcolor{orange!19}70 & \cellcolor{orange!5}3 & \cellcolor{orange!26}80 & \cellcolor{orange!5}20 & \cellcolor{orange!12}57 & \cellcolor{orange!33}83 & \cellcolor{orange!19}73 & \cellcolor{orange!33}83 & \cellcolor{orange!26}77 & \cellcolor{orange!5}0 & \cellcolor{orange!12}57 & \cellcolor{orange!19}67 & \cellcolor{orange!5}3 & \cellcolor{orange!19}70 & \cellcolor{orange!19}63 & \cellcolor{orange!19}63 & \cellcolor{orange!12}43 & \cellcolor{orange!19}63 \\
 & \texttt{Nemo} & \cellcolor{orange!26}73 & \cellcolor{orange!47}93 & \cellcolor{orange!40}87 & \cellcolor{orange!19}73 & \cellcolor{orange!33}83 & \cellcolor{orange!47}97 & \cellcolor{orange!40}90 & \cellcolor{orange!12}57 & \cellcolor{orange!12}53 & \cellcolor{orange!12}57 & \cellcolor{orange!19}67 & \cellcolor{orange!19}67 & \cellcolor{orange!19}73 & \cellcolor{orange!5}10 & \cellcolor{orange!33}80 & \cellcolor{orange!26}80 & \cellcolor{orange!19}63 & \cellcolor{orange!19}67 & \cellcolor{orange!33}83 & \cellcolor{orange!19}73 & \cellcolor{orange!40}87 & \cellcolor{orange!26}73 & \cellcolor{orange!26}73 & \cellcolor{orange!26}77 & \cellcolor{orange!26}80 & \cellcolor{orange!26}77 & \cellcolor{orange!40}87 & \cellcolor{orange!19}73 & \cellcolor{orange!26}77 & \cellcolor{orange!19}73 \\
 & \texttt{Gem4} & \cellcolor{orange!40}90 & \cellcolor{orange!47}93 & \cellcolor{orange!40}87 & \cellcolor{orange!47}93 & \cellcolor{orange!47}97 & \cellcolor{orange!47}97 & \cellcolor{orange!5}17 & \cellcolor{orange!47}93 & \cellcolor{orange!47}93 & \cellcolor{orange!55}100 & \cellcolor{orange!55}100 & \cellcolor{orange!12}53 & \cellcolor{orange!55}100 & \cellcolor{orange!40}90 & \cellcolor{orange!47}97 & \cellcolor{orange!19}73 & \cellcolor{orange!19}73 & \cellcolor{orange!26}80 & \cellcolor{orange!47}97 & \cellcolor{orange!47}97 & \cellcolor{orange!55}100 & \cellcolor{orange!5}0 & \cellcolor{orange!40}90 & \cellcolor{orange!47}93 & \cellcolor{orange!5}0 & \cellcolor{orange!55}100 & \cellcolor{orange!40}87 & \cellcolor{orange!40}90 & \cellcolor{orange!40}90 & \cellcolor{orange!40}87 \\
 & \texttt{Qwen} & \cellcolor{orange!12}60 & \cellcolor{orange!47}93 & \cellcolor{orange!40}87 & \cellcolor{orange!47}97 & \cellcolor{orange!19}70 & \cellcolor{orange!26}80 & \cellcolor{orange!26}73 & \cellcolor{orange!19}70 & \cellcolor{orange!33}83 & \cellcolor{orange!26}80 & \cellcolor{orange!33}83 & \cellcolor{orange!12}60 & \cellcolor{orange!5}17 & \cellcolor{orange!26}79 & \cellcolor{orange!33}83 & \cellcolor{orange!26}80 & \cellcolor{orange!47}93 & \cellcolor{orange!47}93 & \cellcolor{orange!40}87 & \cellcolor{orange!12}47 & \cellcolor{orange!55}100 & \cellcolor{orange!40}90 & \cellcolor{orange!40}90 & \cellcolor{orange!26}77 & \cellcolor{orange!26}80 & \cellcolor{orange!55}100 & \cellcolor{orange!40}87 & \cellcolor{orange!12}37 & \cellcolor{orange!40}87 & \cellcolor{orange!55}100 \\
 & \texttt{Mist} & \cellcolor{orange!19}73 & \cellcolor{orange!26}80 & \cellcolor{orange!47}93 & \cellcolor{orange!26}80 & \cellcolor{orange!19}67 & \cellcolor{orange!19}73 & \cellcolor{orange!40}90 & \cellcolor{orange!47}93 & \cellcolor{orange!19}73 & \cellcolor{orange!26}77 & \cellcolor{orange!33}86 & \cellcolor{orange!12}30 & \cellcolor{orange!33}83 & \cellcolor{orange!19}73 & \cellcolor{orange!26}80 & \cellcolor{orange!5}20 & \cellcolor{orange!19}63 & \cellcolor{orange!19}63 & \cellcolor{orange!26}80 & \cellcolor{orange!26}80 & \cellcolor{orange!19}70 & \cellcolor{orange!26}77 & \cellcolor{orange!19}70 & \cellcolor{orange!33}83 & \cellcolor{orange!40}90 & \cellcolor{orange!19}70 & \cellcolor{orange!40}90 & \cellcolor{orange!40}87 & \cellcolor{orange!26}80 & \cellcolor{orange!40}90 \\
 & \texttt{Aper} & \cellcolor{orange!12}41 & \cellcolor{orange!12}50 & \cellcolor{orange!33}82 & \cellcolor{orange!33}85 & \cellcolor{orange!40}90 & \cellcolor{orange!40}89 & \cellcolor{orange!47}93 & \cellcolor{orange!55}100 & \cellcolor{orange!33}85 & \cellcolor{orange!26}74 & \cellcolor{orange!40}92 & \cellcolor{orange!12}50 & \cellcolor{orange!40}91 & \cellcolor{orange!26}78 & \cellcolor{orange!47}95 & \cellcolor{orange!33}86 & \cellcolor{orange!19}73 & \cellcolor{orange!12}40 & \cellcolor{orange!19}70 & \cellcolor{orange!33}83 & \cellcolor{orange!47}96 & \cellcolor{orange!55}100 & \cellcolor{orange!40}87 & \cellcolor{orange!19}71 & \cellcolor{orange!33}86 & \cellcolor{orange!40}92 & \cellcolor{orange!47}97 & \cellcolor{orange!26}77 & \cellcolor{orange!47}93 & \cellcolor{orange!47}97 \\
\cmidrule(lr){1-32}
\multirow{9}{*}{\texttt{imagegame}} & \texttt{GPT} & \cellcolor{orange!47}100 & \cellcolor{orange!47}99 & \cellcolor{orange!47}100 & \cellcolor{orange!55}100 & \cellcolor{orange!47}100 & \cellcolor{orange!55}100 & \cellcolor{orange!47}100 & \cellcolor{orange!55}100 & \cellcolor{orange!55}100 & \cellcolor{orange!55}100 & \cellcolor{orange!55}100 & \cellcolor{orange!47}98 & \cellcolor{orange!55}100 & \cellcolor{orange!47}99 & \cellcolor{orange!55}100 & \cellcolor{orange!47}100 & \cellcolor{orange!55}100 & \cellcolor{orange!55}100 & \cellcolor{orange!55}100 & \cellcolor{orange!47}100 & \cellcolor{orange!47}100 & \cellcolor{orange!55}100 & \cellcolor{orange!55}100 & \cellcolor{orange!55}100 & \cellcolor{orange!55}100 & \cellcolor{orange!55}100 & \cellcolor{orange!55}100 & \cellcolor{orange!47}100 & \cellcolor{orange!55}100 & \cellcolor{orange!55}100 \\
 & \texttt{Opus} & \cellcolor{orange!19}70 & \cellcolor{orange!40}92 & \cellcolor{orange!19}65 & \cellcolor{orange!33}82 & \cellcolor{orange!47}95 & \cellcolor{orange!19}62 & \cellcolor{orange!12}45 & \cellcolor{orange!19}72 & \cellcolor{orange!47}98 & \cellcolor{orange!40}90 & \cellcolor{orange!33}85 & \cellcolor{orange!40}87 & \cellcolor{orange!47}100 & \cellcolor{orange!47}98 & \cellcolor{orange!12}57 & \cellcolor{orange!40}92 & \cellcolor{orange!47}99 & \cellcolor{orange!40}92 & \cellcolor{orange!12}48 & \cellcolor{orange!47}98 & \cellcolor{orange!40}92 & \cellcolor{orange!26}79 & \cellcolor{orange!26}80 & \cellcolor{orange!19}70 & \cellcolor{orange!47}98 & \cellcolor{orange!47}95 & \cellcolor{orange!55}100 & \cellcolor{orange!5}8 & \cellcolor{orange!40}88 & \cellcolor{orange!5}12 \\
 & \texttt{GLM} & \cellcolor{orange!40}90 & \cellcolor{orange!47}94 & \cellcolor{orange!33}87 & \cellcolor{orange!47}94 & \cellcolor{orange!47}96 & \cellcolor{orange!40}93 & \cellcolor{orange!12}38 & \cellcolor{orange!47}94 & \cellcolor{orange!40}91 & \cellcolor{orange!47}96 & \cellcolor{orange!47}95 & \cellcolor{orange!12}46 & \cellcolor{orange!47}98 & \cellcolor{orange!47}95 & \cellcolor{orange!26}77 & \cellcolor{orange!19}65 & \cellcolor{orange!33}86 & \cellcolor{orange!12}52 & \cellcolor{orange!47}96 & \cellcolor{orange!40}92 & \cellcolor{orange!40}87 & \cellcolor{orange!40}93 & \cellcolor{orange!47}98 & \cellcolor{orange!40}91 & \cellcolor{orange!47}96 & \cellcolor{orange!47}95 & \cellcolor{orange!47}97 & \cellcolor{orange!19}61 & \cellcolor{orange!40}93 & \cellcolor{orange!5}18 \\
 & \texttt{DSV4} & \cellcolor{orange!33}83 & \cellcolor{orange!33}82 & \cellcolor{orange!26}74 & \cellcolor{orange!26}80 & \cellcolor{orange!40}92 & \cellcolor{orange!19}67 & \cellcolor{orange!12}38 & \cellcolor{orange!19}71 & \cellcolor{orange!33}82 & \cellcolor{orange!33}82 & \cellcolor{orange!33}81 & \cellcolor{orange!5}0 & \cellcolor{orange!19}64 & \cellcolor{orange!26}74 & \cellcolor{orange!12}60 & \cellcolor{orange!5}0 & \cellcolor{orange!19}71 & \cellcolor{orange!19}64 & \cellcolor{orange!33}81 & \cellcolor{orange!33}81 & \cellcolor{orange!26}78 & \cellcolor{orange!26}78 & \cellcolor{orange!33}83 & \cellcolor{orange!19}61 & \cellcolor{orange!19}65 & \cellcolor{orange!26}74 & \cellcolor{orange!19}67 & \cellcolor{orange!12}61 & \cellcolor{orange!40}87 & \cellcolor{orange!12}33 \\
 & \texttt{Nemo} & \cellcolor{orange!12}53 & \cellcolor{orange!26}73 & \cellcolor{orange!12}48 & \cellcolor{orange!40}90 & \cellcolor{orange!47}94 & \cellcolor{orange!26}79 & \cellcolor{orange!26}78 & \cellcolor{orange!33}86 & \cellcolor{orange!26}76 & \cellcolor{orange!26}74 & \cellcolor{orange!40}92 & \cellcolor{orange!5}8 & \cellcolor{orange!33}86 & \cellcolor{orange!26}75 & \cellcolor{orange!5}21 & \cellcolor{orange!12}26 & \cellcolor{orange!12}58 & \cellcolor{orange!12}42 & \cellcolor{orange!26}76 & \cellcolor{orange!33}80 & \cellcolor{orange!33}86 & \cellcolor{orange!40}93 & \cellcolor{orange!47}93 & \cellcolor{orange!12}36 & \cellcolor{orange!12}45 & \cellcolor{orange!26}75 & \cellcolor{orange!40}91 & \cellcolor{orange!5}23 & \cellcolor{orange!33}85 & \cellcolor{orange!12}33 \\
 & \texttt{Gem4} & \cellcolor{orange!5}18 & \cellcolor{orange!40}89 & \cellcolor{orange!12}45 & \cellcolor{orange!33}82 & \cellcolor{orange!33}84 & \cellcolor{orange!19}66 & \cellcolor{orange!5}7 & \cellcolor{orange!26}75 & \cellcolor{orange!19}66 & \cellcolor{orange!33}85 & \cellcolor{orange!19}62 & \cellcolor{orange!5}12 & \cellcolor{orange!33}85 & \cellcolor{orange!26}78 & \cellcolor{orange!19}61 & \cellcolor{orange!5}0 & \cellcolor{orange!26}75 & \cellcolor{orange!12}29 & \cellcolor{orange!19}69 & \cellcolor{orange!12}61 & \cellcolor{orange!33}81 & \cellcolor{orange!19}65 & \cellcolor{orange!26}79 & \cellcolor{orange!12}54 & \cellcolor{orange!19}66 & \cellcolor{orange!26}78 & \cellcolor{orange!19}69 & \cellcolor{orange!5}0 & \cellcolor{orange!33}82 & \cellcolor{orange!5}0 \\
 & \texttt{Qwen} & \cellcolor{orange!33}83 & \cellcolor{orange!40}93 & \cellcolor{orange!47}94 & \cellcolor{orange!33}84 & \cellcolor{orange!47}95 & \cellcolor{orange!47}98 & \cellcolor{orange!47}95 & \cellcolor{orange!33}84 & \cellcolor{orange!40}92 & \cellcolor{orange!12}39 & \cellcolor{orange!26}77 & \cellcolor{orange!12}57 & \cellcolor{orange!33}81 & \cellcolor{orange!19}72 & \cellcolor{orange!26}80 & \cellcolor{orange!12}26 & \cellcolor{orange!40}88 & \cellcolor{orange!33}82 & \cellcolor{orange!26}74 & \cellcolor{orange!19}61 & \cellcolor{orange!40}91 & \cellcolor{orange!19}66 & \cellcolor{orange!33}86 & \cellcolor{orange!47}98 & \cellcolor{orange!40}91 & \cellcolor{orange!33}83 & \cellcolor{orange!40}90 & \cellcolor{orange!40}88 & \cellcolor{orange!40}92 & \cellcolor{orange!5}5 \\
 & \texttt{Mist} & \cellcolor{orange!12}49 & \cellcolor{orange!12}53 & \cellcolor{orange!5}9 & \cellcolor{orange!12}51 & \cellcolor{orange!19}68 & \cellcolor{orange!12}47 & \cellcolor{orange!5}20 & \cellcolor{orange!19}63 & \cellcolor{orange!12}49 & \cellcolor{orange!12}58 & \cellcolor{orange!40}88 & \cellcolor{orange!5}0 & \cellcolor{orange!12}39 & \cellcolor{orange!12}28 & \cellcolor{orange!5}0 & \cellcolor{orange!5}0 & \cellcolor{orange!12}42 & \cellcolor{orange!5}8 & \cellcolor{orange!26}77 & \cellcolor{orange!12}58 & \cellcolor{orange!12}52 & \cellcolor{orange!12}36 & \cellcolor{orange!12}45 & \cellcolor{orange!5}8 & \cellcolor{orange!5}12 & \cellcolor{orange!12}36 & \cellcolor{orange!12}35 & \cellcolor{orange!5}5 & \cellcolor{orange!19}65 & \cellcolor{orange!5}12 \\
 & \texttt{Aper} & \cellcolor{orange!12}25 & \cellcolor{orange!5}8 & \cellcolor{orange!5}8 & \cellcolor{orange!5}15 & \cellcolor{orange!5}4 & \cellcolor{orange!5}11 & \cellcolor{orange!5}9 & \cellcolor{orange!5}13 & \cellcolor{orange!5}8 & \cellcolor{orange!5}9 & \cellcolor{orange!5}14 & \cellcolor{orange!5}5 & \cellcolor{orange!5}10 & \cellcolor{orange!5}12 & \cellcolor{orange!5}10 & \cellcolor{orange!5}7 & \cellcolor{orange!5}8 & \cellcolor{orange!5}6 & \cellcolor{orange!5}16 & \cellcolor{orange!5}6 & \cellcolor{orange!5}2 & \cellcolor{orange!5}9 & \cellcolor{orange!5}9 & \cellcolor{orange!5}6 & \cellcolor{orange!5}6 & \cellcolor{orange!5}13 & \cellcolor{orange!5}17 & \cellcolor{orange!5}0 & \cellcolor{orange!5}14 & \cellcolor{orange!5}4 \\
\bottomrule
\end{tabular}%
}
\caption{clemscore for the \textbf{Grounded reference \& spatial language} capability cluster: one row per game and model, one column per language. Shading uses this cluster's own score distribution (equal-population octiles, darker = higher); ``-'' marks a missing run. Models: \texttt{GPT}~=~GPT-5.4; \texttt{Opus}~=~Claude Opus 4.8; \texttt{GLM}~=~GLM-5.2; \texttt{DSV4}~=~DeepSeek-V4-Pro; \texttt{Nemo}~=~Nemotron-3-Ultra; \texttt{Gem4}~=~Gemma-4-26B; \texttt{Qwen}~=~Qwen3.6-35B-A3B; \texttt{Mist}~=~Mistral Large 3; \texttt{Aper}~=~Apertus-v1.5-8B.}
\label{tab:cgmT-grref}
\end{table*}

\begin{table*}[t]
\centering
\scriptsize
\setlength{\tabcolsep}{2.2pt}
\renewcommand{\arraystretch}{1.05}
\resizebox{\textwidth}{!}{%
\begin{tabular}{@{}ll *{30}{c}@{}}
\toprule
\textbf{Game} & \textbf{Model} & \textbf{ar} & \textbf{bg} & \textbf{cs} & \textbf{da} & \textbf{de} & \textbf{el} & \textbf{en} & \textbf{es} & \textbf{et} & \textbf{fi} & \textbf{fr} & \textbf{ga} & \textbf{hr} & \textbf{hu} & \textbf{it} & \textbf{lt} & \textbf{lv} & \textbf{mt} & \textbf{nl} & \textbf{pl} & \textbf{pt} & \textbf{ro} & \textbf{ru} & \textbf{sk} & \textbf{sl} & \textbf{sr} & \textbf{sv} & \textbf{tr} & \textbf{uk} & \textbf{zh} \\
\midrule
\multirow{9}{*}{\texttt{matchit}} & \texttt{GPT} & \cellcolor{violet!55}100 & \cellcolor{violet!55}100 & \cellcolor{violet!55}100 & \cellcolor{violet!55}100 & \cellcolor{violet!55}100 & \cellcolor{violet!55}100 & \cellcolor{violet!55}100 & \cellcolor{violet!55}95 & \cellcolor{violet!55}100 & \cellcolor{violet!55}95 & \cellcolor{violet!55}100 & \cellcolor{violet!55}100 & \cellcolor{violet!55}100 & \cellcolor{violet!55}100 & \cellcolor{violet!55}100 & \cellcolor{violet!55}95 & \cellcolor{violet!55}100 & \cellcolor{violet!55}100 & \cellcolor{violet!55}100 & \cellcolor{violet!55}95 & \cellcolor{violet!55}95 & \cellcolor{violet!55}100 & \cellcolor{violet!55}100 & \cellcolor{violet!55}100 & \cellcolor{violet!55}100 & \cellcolor{violet!55}100 & \cellcolor{violet!55}100 & \cellcolor{violet!55}100 & \cellcolor{violet!55}100 & \cellcolor{violet!55}95 \\
\bottomrule
\end{tabular}%
}
\caption{clemscore for the \textbf{Discourse \& grounding} capability cluster: one row per game and model, one column per language. Shading uses this cluster's own score distribution (equal-population octiles, darker = higher); ``-'' marks a missing run. Models: \texttt{GPT}~=~GPT-5.4; \texttt{Opus}~=~Claude Opus 4.8; \texttt{GLM}~=~GLM-5.2; \texttt{DSV4}~=~DeepSeek-V4-Pro; \texttt{Nemo}~=~Nemotron-3-Ultra; \texttt{Gem4}~=~Gemma-4-26B; \texttt{Qwen}~=~Qwen3.6-35B-A3B; \texttt{Mist}~=~Mistral Large 3; \texttt{Aper}~=~Apertus-v1.5-8B.}
\label{tab:cgmT-discourse}
\end{table*}

\begin{table*}[t]
\centering
\scriptsize
\setlength{\tabcolsep}{2.2pt}
\renewcommand{\arraystretch}{1.05}
\resizebox{\textwidth}{!}{%
\begin{tabular}{@{}ll *{30}{c}@{}}
\toprule
\textbf{Game} & \textbf{Model} & \textbf{ar} & \textbf{bg} & \textbf{cs} & \textbf{da} & \textbf{de} & \textbf{el} & \textbf{en} & \textbf{es} & \textbf{et} & \textbf{fi} & \textbf{fr} & \textbf{ga} & \textbf{hr} & \textbf{hu} & \textbf{it} & \textbf{lt} & \textbf{lv} & \textbf{mt} & \textbf{nl} & \textbf{pl} & \textbf{pt} & \textbf{ro} & \textbf{ru} & \textbf{sk} & \textbf{sl} & \textbf{sr} & \textbf{sv} & \textbf{tr} & \textbf{uk} & \textbf{zh} \\
\midrule
\multirow{9}{*}{\texttt{dond}} & \texttt{GPT} & \cellcolor{olive!55}99 & \cellcolor{olive!47}97 & \cellcolor{olive!47}98 & \cellcolor{olive!47}97 & \cellcolor{olive!55}100 & \cellcolor{olive!55}100 & \cellcolor{olive!55}99 & \cellcolor{olive!55}100 & \cellcolor{olive!55}100 & \cellcolor{olive!55}100 & \cellcolor{olive!55}100 & \cellcolor{olive!55}100 & \cellcolor{olive!55}100 & \cellcolor{olive!47}98 & \cellcolor{olive!55}100 & \cellcolor{olive!47}88 & \cellcolor{olive!47}98 & \cellcolor{olive!55}99 & \cellcolor{olive!55}100 & \cellcolor{olive!55}99 & \cellcolor{olive!55}100 & \cellcolor{olive!55}100 & \cellcolor{olive!47}90 & \cellcolor{olive!47}98 & \cellcolor{olive!55}100 & \cellcolor{olive!47}89 & \cellcolor{olive!55}100 & \cellcolor{olive!55}100 & \cellcolor{olive!47}99 & \cellcolor{olive!47}99 \\
\bottomrule
\end{tabular}%
}
\caption{clemscore for the \textbf{Strategic / social reasoning} capability cluster: one row per game and model, one column per language. Shading uses this cluster's own score distribution (equal-population octiles, darker = higher); ``-'' marks a missing run. Models: \texttt{GPT}~=~GPT-5.4; \texttt{Opus}~=~Claude Opus 4.8; \texttt{GLM}~=~GLM-5.2; \texttt{DSV4}~=~DeepSeek-V4-Pro; \texttt{Nemo}~=~Nemotron-3-Ultra; \texttt{Gem4}~=~Gemma-4-26B; \texttt{Qwen}~=~Qwen3.6-35B-A3B; \texttt{Mist}~=~Mistral Large 3; \texttt{Aper}~=~Apertus-v1.5-8B.}
\label{tab:cgmT-strategic}
\end{table*}

\begin{table*}[t]
\centering
\scriptsize
\setlength{\tabcolsep}{2.2pt}
\renewcommand{\arraystretch}{1.05}
\resizebox{\textwidth}{!}{%
\begin{tabular}{@{}ll *{30}{c}@{}}
\toprule
\textbf{Game} & \textbf{Model} & \textbf{ar} & \textbf{bg} & \textbf{cs} & \textbf{da} & \textbf{de} & \textbf{el} & \textbf{en} & \textbf{es} & \textbf{et} & \textbf{fi} & \textbf{fr} & \textbf{ga} & \textbf{hr} & \textbf{hu} & \textbf{it} & \textbf{lt} & \textbf{lv} & \textbf{mt} & \textbf{nl} & \textbf{pl} & \textbf{pt} & \textbf{ro} & \textbf{ru} & \textbf{sk} & \textbf{sl} & \textbf{sr} & \textbf{sv} & \textbf{tr} & \textbf{uk} & \textbf{zh} \\
\midrule
\multirow{9}{*}{\texttt{tmw}} & \texttt{GPT} & \cellcolor{cyan!47}91 & \cellcolor{cyan!47}92 & \cellcolor{cyan!47}90 & \cellcolor{cyan!47}91 & \cellcolor{cyan!47}90 & \cellcolor{cyan!47}92 & \cellcolor{cyan!47}91 & \cellcolor{cyan!47}89 & \cellcolor{cyan!47}92 & \cellcolor{cyan!47}91 & \cellcolor{cyan!47}91 & \cellcolor{cyan!47}92 & \cellcolor{cyan!47}90 & \cellcolor{cyan!47}90 & \cellcolor{cyan!47}90 & \cellcolor{cyan!47}90 & \cellcolor{cyan!47}90 & \cellcolor{cyan!47}91 & \cellcolor{cyan!47}89 & \cellcolor{cyan!47}90 & \cellcolor{cyan!47}90 & \cellcolor{cyan!47}92 & \cellcolor{cyan!47}91 & \cellcolor{cyan!47}91 & \cellcolor{cyan!47}90 & \cellcolor{cyan!47}89 & \cellcolor{cyan!47}90 & \cellcolor{cyan!47}93 & \cellcolor{cyan!47}91 & \cellcolor{cyan!47}89 \\
 & \texttt{Opus} & \cellcolor{cyan!40}86 & \cellcolor{cyan!40}86 & \cellcolor{cyan!47}89 & \cellcolor{cyan!47}89 & \cellcolor{cyan!40}84 & \cellcolor{cyan!40}86 & \cellcolor{cyan!40}83 & \cellcolor{cyan!40}84 & \cellcolor{cyan!40}86 & \cellcolor{cyan!40}84 & \cellcolor{cyan!47}88 & \cellcolor{cyan!19}62 & \cellcolor{cyan!40}86 & \cellcolor{cyan!40}85 & \cellcolor{cyan!40}87 & \cellcolor{cyan!40}81 & \cellcolor{cyan!33}78 & \cellcolor{cyan!33}76 & \cellcolor{cyan!40}85 & \cellcolor{cyan!40}88 & \cellcolor{cyan!40}86 & \cellcolor{cyan!40}87 & \cellcolor{cyan!40}84 & \cellcolor{cyan!40}85 & \cellcolor{cyan!40}87 & \cellcolor{cyan!40}86 & \cellcolor{cyan!47}90 & \cellcolor{cyan!40}85 & \cellcolor{cyan!40}83 & \cellcolor{cyan!40}84 \\
\bottomrule
\end{tabular}%
}
\caption{clemscore for the \textbf{Spatial reasoning \& planning} capability cluster: one row per game and model, one column per language. Shading uses this cluster's own score distribution (equal-population octiles, darker = higher); ``-'' marks a missing run. Models: \texttt{GPT}~=~GPT-5.4; \texttt{Opus}~=~Claude Opus 4.8; \texttt{GLM}~=~GLM-5.2; \texttt{DSV4}~=~DeepSeek-V4-Pro; \texttt{Nemo}~=~Nemotron-3-Ultra; \texttt{Gem4}~=~Gemma-4-26B; \texttt{Qwen}~=~Qwen3.6-35B-A3B; \texttt{Mist}~=~Mistral Large 3; \texttt{Aper}~=~Apertus-v1.5-8B.}
\label{tab:cgmT-spatial}
\end{table*}

\section{Token Fertility Calculation}
\label{app:fertility}
\label{sec:token-fertility}

Token fertility is measured on natively written text rather than on game transcripts. For each language we collect Wikipedia articles about the countries where it is spoken (the same country lists used for LEF; for widely spoken languages, the ten largest by GDP), retrieved from that language's own Wikipedia via the MediaWiki API and cropped to the first 12{,}000 characters per article so that an identical corpus fits into a single request for every model. Fertility is total tokens divided by total words, with words whitespace-delimited except in Chinese, where we count non-whitespace characters.

Token counts use each model's own tokeniser: the Hugging Face \texttt{AutoTokenizer} for most open-weight models, \texttt{tiktoken} (\texttt{o200k\_base}) for GPT-5.4, and Mistral's \texttt{mistral-common} (Tekken) tokeniser for Mistral-Large-3, Nemotron-3-Ultra and Apertus, which all share it and therefore have identical fertility values. Claude Opus 4.8 has no public tokeniser; its counts come from the Anthropic Messages API \texttt{count\_tokens} endpoint, after subtracting the constant per-message overhead measured with a one-character probe. Anthropic documents these as close estimates of billed usage and notes that Opus~4.7 and later use a tokeniser producing roughly 30\% more tokens than earlier Claude models for the same text, which is a different comparison from the gap to the median model reported in Section~\ref{sec:lef}. That gap averages 50\% across the EU-24 but varies widely by language, from $+11\%$ for Bulgarian to $+92\%$ for German.

\section{Linguistic Economic Footprint: Data}
\label{app:lef}

\begin{figure*}[t]
    \centering
    \includegraphics[width=1.0\linewidth]{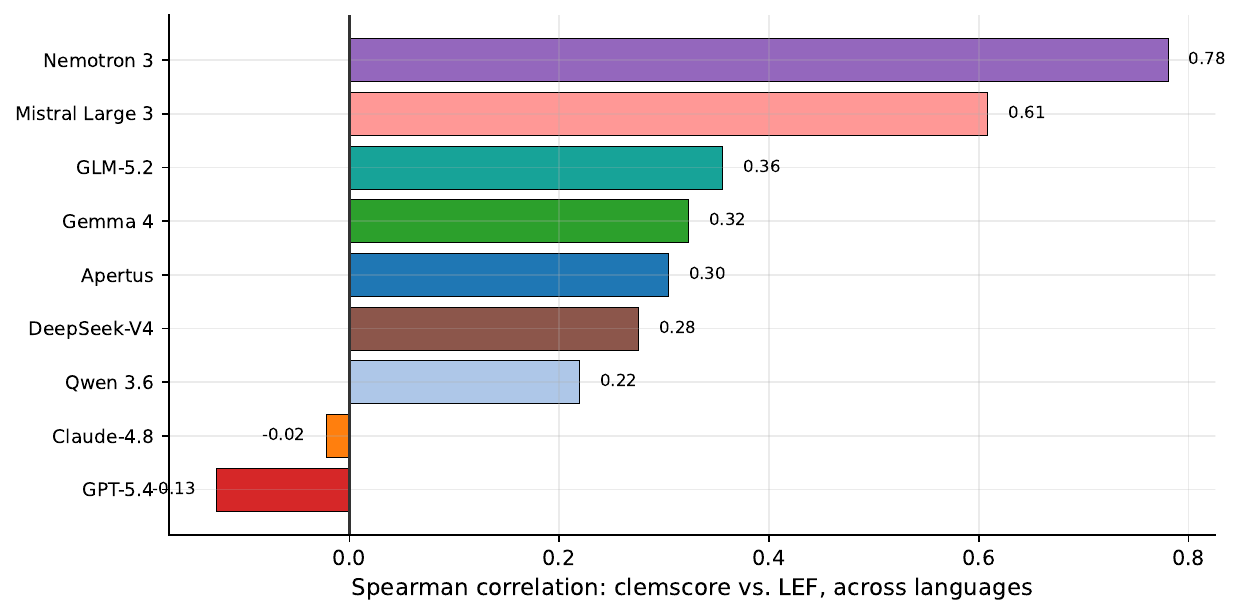}
    \caption{Spearman correlation between each model's clemscore and LEF across the 30 benchmark languages. Higher values indicate performance more skewed towards economically dominant languages; values near zero indicate performance largely independent of a language's economic footprint.}
    \label{fig:lef-corr}
\end{figure*}

Figure~\ref{fig:lef-corr} gives the per-model clemscore--LEF correlations discussed in Section~\ref{sec:lef}. Table~\ref{tab:lef_countries} gives the underlying data, one row per language. \textbf{Countries} lists every country where the language is official or co-official, as \textsc{code}~(population in millions, 2024 nominal GDP in bn\,USD from the World Bank \texttt{NY.GDP.MKTP.CD} series, share of the population speaking it). Shares use Ethnologue total users (29th ed., 2024), which count first- and second-language speakers where the source reports both, capped at 1.0 where speaker and population estimates come from different sources. \textbf{Spk} sums speakers over the listed countries, and \textbf{LEF} is the speaker-weighted footprint $\sum_c \mathrm{GDP}(c)\times\min(1,\mathrm{speakers}/\mathrm{population})$ in bn\,USD.

% Table generated by paper/lef/m_gdlp.py; regenerate with `python3 paper/lef/m_gdlp.py`
% AUTO-GENERATED by paper/lef/m_gdlp.py -- do not edit by hand.
\begin{table*}[t]
\centering
\scriptsize
\setlength{\tabcolsep}{3pt}
\renewcommand{\arraystretch}{1.25}
\begin{tabularx}{\textwidth}{@{}ll X rr@{}}
\toprule
\textbf{Code} & \textbf{Language} &
  \textbf{Countries (pop.\,M,\ GDP\,bn\,USD,\ share\,\%)} &
  \textbf{Spk\,(M)} & \textbf{LEF\,(bn\,USD)} \\
\midrule
\multicolumn{5}{@{}l}{\rule{0pt}{9pt}\textit{Germanic}} \\[-2pt]
\texttt{da} & Danish & \mbox{DK~(6,\,424,\,94.0\%)} & 5.64 & 399 \\
\texttt{de} & German & \mbox{DE~(83.5,\,4,686,\,96.7\%)}, \mbox{AT~(9.2,\,535,\,96.7\%)}, \mbox{CH~(9,\,937,\,61.1\%)}, \mbox{LU~(0.68,\,93.3,\,66.8\%)}, \mbox{BE~(11.9,\,671,\,0.7\%)}, \mbox{LI~(0.04,\,7,\,95.0\%)} & 95.7 & 5,693 \\
\texttt{en} & English & \mbox{US~(340,\,28,751,\,87.9\%)}, \mbox{GB~(69.2,\,3,686,\,92.3\%)}, \mbox{CA~(41.3,\,2,244,\,76.5\%)}, \mbox{AU~(27.2,\,1,757,\,89.7\%)}, \mbox{IE~(5.4,\,609,\,87.5\%)}, \mbox{NZ~(5.3,\,260,\,93.3\%)}, \mbox{IN~(1,451,\,3,910,\,19.2\%)}, \mbox{ZA~(64,\,401,\,31.2\%)}, \mbox{NG~(233,\,252,\,25.8\%)}, \mbox{SG~(6,\,547,\,51.2\%)}, \mbox{PH~(116,\,462,\,46.3\%)}, \mbox{KE~(56.4,\,120,\,78.1\%)}, \mbox{GH~(34.4,\,82.3,\,41.3\%)}, \mbox{TZ~(68.6,\,78.8,\,9.8\%)}, \mbox{UG~(50,\,53.9,\,58.0\%)}, \mbox{ZM~(21.3,\,25.3,\,16.0\%)}, \mbox{ZW~(16.6,\,41.5,\,31.1\%)}, \mbox{BW~(2.5,\,19.4,\,42.8\%)}, \mbox{NA~(3,\,13.4,\,14.8\%)}, \mbox{RW~(14.3,\,14.3,\,13.9\%)}, \mbox{MT~(0.57,\,25,\,72.1\%)}, \mbox{PK~(251,\,372,\,6.8\%)} & 968 & 34,425 \\
\texttt{nl} & Dutch & \mbox{NL~(18,\,1,215,\,91.1\%)}, \mbox{BE~(11.9,\,671,\,66.4\%)}, \mbox{SR~(0.63,\,4.4,\,21.0\%)} & 24.4 & 1,554 \\
\texttt{sv} & Swedish & \mbox{SE~(10.6,\,604,\,93.2\%)}, \mbox{FI~(5.6,\,299,\,5.2\%)} & 10.2 & 578 \\
\multicolumn{5}{@{}l}{\rule{0pt}{9pt}\textit{Romance}} \\[-2pt]
\texttt{es} & Spanish & \mbox{ES~(48.8,\,1,726,\,96.2\%)}, \mbox{MX~(131,\,1,856,\,97.2\%)}, \mbox{CO~(52.9,\,419,\,97.9\%)}, \mbox{AR~(45.7,\,638,\,100.0\%)}, \mbox{CL~(19.8,\,330,\,100.0\%)}, \mbox{PE~(34.2,\,289,\,97.2\%)}, \mbox{VE~(28.4,\,120,\,100.0\%)}, \mbox{EC~(18.1,\,125,\,97.4\%)}, \mbox{GT~(18.4,\,113,\,89.8\%)}, \mbox{CU~(11,\,107,\,100.0\%)}, \mbox{BO~(12.4,\,54.9,\,82.1\%)}, \mbox{DO~(11.4,\,124,\,85.8\%)}, \mbox{HN~(10.8,\,37.1,\,95.3\%)}, \mbox{PY~(6.9,\,44.5,\,94.9\%)}, \mbox{SV~(6.3,\,35.4,\,100.0\%)}, \mbox{NI~(6.9,\,19.7,\,97.2\%)}, \mbox{CR~(5.1,\,95.4,\,100.0\%)}, \mbox{PA~(4.5,\,86.5,\,97.4\%)}, \mbox{UY~(3.4,\,81,\,100.0\%)}, \mbox{GQ~(1.9,\,12.8,\,64.5\%)} & 465 & 6,126 \\
\texttt{fr} & French & \mbox{FR~(68.6,\,3,160,\,96.3\%)}, \mbox{BE~(11.9,\,671,\,82.1\%)}, \mbox{CH~(9,\,937,\,66.2\%)}, \mbox{CA~(41.3,\,2,244,\,27.3\%)}, \mbox{LU~(0.68,\,93.3,\,87.6\%)}, \mbox{CD~(109,\,71,\,48.3\%)}, \mbox{CI~(31.9,\,87.1,\,35.9\%)}, \mbox{CM~(29.1,\,53.3,\,40.9\%)}, \mbox{SN~(18.5,\,32.8,\,27.4\%)}, \mbox{ML~(24.5,\,26.8,\,19.4\%)}, \mbox{BF~(23.5,\,23.1,\,21.5\%)}, \mbox{NE~(27,\,19.9,\,14.0\%)}, \mbox{TD~(20.3,\,19.5,\,11.8\%)}, \mbox{GN~(14.8,\,25,\,27.0\%)}, \mbox{RW~(14.3,\,14.3,\,5.2\%)}, \mbox{BJ~(14.5,\,21.5,\,32.6\%)}, \mbox{BI~(14,\,3.1,\,8.5\%)}, \mbox{TG~(9.5,\,10.7,\,40.2\%)}, \mbox{CF~(5.3,\,2.8,\,27.3\%)}, \mbox{GA~(2.5,\,20.9,\,65.4\%)}, \mbox{CG~(6.3,\,15.7,\,60.4\%)}, \mbox{MG~(32,\,17.4,\,25.9\%)}, \mbox{KM~(0.87,\,1.4,\,34.6\%)}, \mbox{DJ~(1.2,\,4.2,\,50.1\%)}, \mbox{HT~(11.8,\,25.2,\,4.2\%)}, \mbox{MU~(1.2,\,14.9,\,76.7\%)}, \mbox{SC~(0.12,\,2.2,\,48.5\%)}, \mbox{VU~(0.33,\,1.1,\,25.1\%)} & 223 & 5,086 \\
\texttt{it} & Italian & \mbox{IT~(59,\,2,381,\,94.1\%)}, \mbox{CH~(9,\,937,\,7.5\%)}, \mbox{SM~(0.034,\,2.1,\,100.0\%)}, \mbox{VA~(0.001,\,0.5,\,100.0\%)} & 56.2 & 2,314 \\
\texttt{pt} & Portuguese & \mbox{BR~(212,\,2,186,\,100.0\%)}, \mbox{PT~(10.7,\,313,\,93.6\%)}, \mbox{AO~(37.9,\,101,\,58.2\%)}, \mbox{MZ~(34.6,\,22.7,\,42.4\%)}, \mbox{GW~(2.2,\,2.2,\,18.0\%)}, \mbox{CV~(0.52,\,2.7,\,70.7\%)}, \mbox{ST~(0.24,\,0.8,\,95.0\%)}, \mbox{TL~(1.4,\,1.9,\,35.7\%)}, \mbox{GQ~(1.9,\,12.8,\,0.3\%)}, \mbox{MO~(0.69,\,49.5,\,2.8\%)} & 264 & 2,553 \\
\texttt{ro} & Romanian & \mbox{RO~(19.1,\,383,\,79.6\%)}, \mbox{MD~(2.4,\,18.2,\,100.0\%)} & 17.7 & 323 \\
\multicolumn{5}{@{}l}{\rule{0pt}{9pt}\textit{Slavic}} \\[-2pt]
\texttt{bg} & Bulgarian & \mbox{BG~(6.4,\,113,\,89.4\%)} & 5.723 & 101 \\
\texttt{cs} & Czech & \mbox{CZ~(10.9,\,347,\,86.6\%)} & 9.443 & 301 \\
\texttt{hr} & Croatian & \mbox{HR~(3.9,\,93,\,94.4\%)}, \mbox{BA~(3.2,\,29.6,\,14.6\%)} & 4.149 & 92.1 \\
\texttt{pl} & Polish & \mbox{PL~(36.6,\,918,\,100.0\%)} & 39.5 & 918 \\
\texttt{ru} & Russian & \mbox{RU~(144,\,2,174,\,93.4\%)}, \mbox{BY~(9.1,\,76,\,77.2\%)}, \mbox{KZ~(20.6,\,292,\,87.1\%)}, \mbox{KG~(7.2,\,17.5,\,50.9\%)} & 163 & 2,351 \\
\texttt{sk} & Slovak & \mbox{SK~(5.4,\,141,\,90.3\%)} & 4.877 & 127 \\
\texttt{sl} & Slovenian & \mbox{SI~(2.1,\,73,\,98.1\%)} & 2.0606 & 71.6 \\
\texttt{sr} & Serbian & \mbox{RS~(6.6,\,90.1,\,91.7\%)}, \mbox{BA~(3.2,\,29.6,\,33.6\%)}, \mbox{ME~(0.62,\,8.3,\,51.5\%)}, \mbox{XK~(1.6,\,11.2,\,6.2\%)} & 7.545 & 97.5 \\
\texttt{uk} & Ukrainian & \mbox{UA~(37.9,\,191,\,58.6\%)} & 22.2 & 112 \\
\multicolumn{5}{@{}l}{\rule{0pt}{9pt}\textit{Baltic}} \\[-2pt]
\texttt{lt} & Lithuanian & \mbox{LT~(2.9,\,84.9,\,92.9\%)} & 2.695 & 78.9 \\
\texttt{lv} & Latvian & \mbox{LV~(1.9,\,43.7,\,96.8\%)} & 1.84 & 42.3 \\
\multicolumn{5}{@{}l}{\rule{0pt}{9pt}\textit{Uralic}} \\[-2pt]
\texttt{et} & Estonian & \mbox{EE~(1.4,\,43.1,\,63.9\%)} & 0.895 & 27.6 \\
\texttt{fi} & Finnish & \mbox{FI~(5.6,\,299,\,94.6\%)} & 5.3 & 283 \\
\texttt{hu} & Hungarian & \mbox{HU~(9.6,\,223,\,99.3\%)} & 9.5376 & 221 \\
\multicolumn{5}{@{}l}{\rule{0pt}{9pt}\textit{Celtic}} \\[-2pt]
\texttt{ga} & Irish & \mbox{IE~(5.4,\,609,\,21.7\%)} & 1.171 & 132 \\
\multicolumn{5}{@{}l}{\rule{0pt}{9pt}\textit{Hellenic}} \\[-2pt]
\texttt{el} & Greek & \mbox{GR~(10.4,\,256,\,98.0\%)}, \mbox{CY~(1.4,\,37.6,\,84.9\%)} & 11.4 & 283 \\
\multicolumn{5}{@{}l}{\rule{0pt}{9pt}\textit{Semitic}} \\[-2pt]
\texttt{ar} & Arabic & \mbox{SA~(35.3,\,1,240,\,81.9\%)}, \mbox{AE~(11,\,552,\,31.7\%)}, \mbox{EG~(116,\,389,\,66.3\%)}, \mbox{IQ~(46,\,280,\,76.3\%)}, \mbox{DZ~(46.8,\,269,\,75.6\%)}, \mbox{QA~(2.9,\,219,\,12.1\%)}, \mbox{KW~(4.9,\,160,\,76.1\%)}, \mbox{MA~(38.1,\,161,\,71.7\%)}, \mbox{OM~(5.3,\,107,\,42.5\%)}, \mbox{JO~(11.6,\,53.4,\,74.0\%)}, \mbox{TN~(12.3,\,51.3,\,79.6\%)}, \mbox{BH~(1.6,\,47.1,\,51.1\%)}, \mbox{LY~(7.4,\,48.5,\,84.5\%)}, \mbox{YE~(40.6,\,22,\,52.5\%)}, \mbox{SY~(24.7,\,60,\,81.8\%)}, \mbox{LB~(5.8,\,21,\,82.4\%)}, \mbox{SD~(50.4,\,49.7,\,53.8\%)}, \mbox{MR~(5.2,\,10.9,\,48.3\%)}, \mbox{SO~(19,\,12,\,2.6\%)}, \mbox{DJ~(1.2,\,4.2,\,25.8\%)}, \mbox{KM~(0.87,\,1.4,\,34.5\%)}, \mbox{PS~(5.3,\,13.7,\,95.7\%)} & 321 & 2,444 \\
\texttt{mt} & Maltese & \mbox{MT~(0.57,\,25,\,88.4\%)} & 0.504 & 22.1 \\
\multicolumn{5}{@{}l}{\rule{0pt}{9pt}\textit{Turkic}} \\[-2pt]
\texttt{tr} & Turkish & \mbox{TR~(85.5,\,1,359,\,99.1\%)}, \mbox{CY~(1.4,\,37.6,\,21.4\%)} & 85 & 1,355 \\
\multicolumn{5}{@{}l}{\rule{0pt}{9pt}\textit{Sino-Tibetan}} \\[-2pt]
\texttt{zh} & Chinese & \mbox{CN~(1,409,\,18,744,\,80.8\%)}, \mbox{TW~(23.4,\,775,\,89.9\%)}, \mbox{SG~(6,\,547,\,47.3\%)}, \mbox{MO~(0.69,\,49.5,\,83.0\%)}, \mbox{HK~(7.5,\,407,\,89.7\%)} & 1,169 & 16,501 \\
\bottomrule
\end{tabularx}
\caption{Countries, speaker estimates, and computed LEF per language. Country codes are ISO 3166-1 alpha-2 (XK = Kosovo (EU practical code; non-ISO), PS = Palestine). Speaker counts are total users from Ethnologue (29th ed., 2024), including L1+L2 where available.}
\label{tab:lef_countries}
\end{table*}

\section{Web Data Availability per Language}
\label{app:data-availability}

\begin{figure*}[t]
    \centering
    \includegraphics[width=1.0\linewidth]{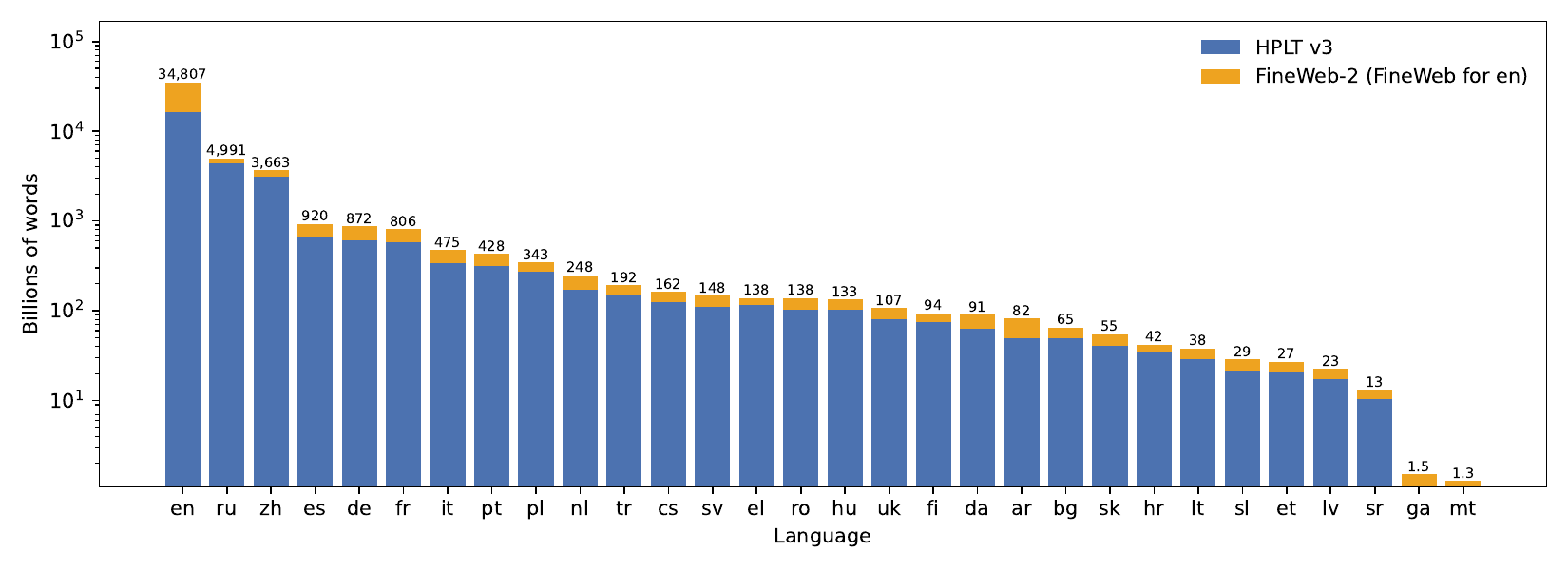}
    \caption{Web data availability per language (billions of words, log scale) in HPLT~v3 (blue) and FineWeb-2 (orange; FineWeb for English), sorted by combined total. Word counts are as reported by each dataset.}
    \label{fig:lang-words}
\end{figure*}

Figure~\ref{fig:lang-words} shows the web-crawled text available for the 30 benchmark languages in two large multilingual pretraining corpora: HPLT~v3\footnote{\url{https://hplt-project.org/datasets/v3.0}} and FineWeb-2\footnote{\url{https://huggingface.co/datasets/HuggingFaceFW/fineweb-2}} (FineWeb for English, as FineWeb-2 covers only non-English languages). Counts are as reported by each dataset and are therefore approximate rather than strictly comparable across the two. The distribution is highly skewed: English alone accounts for roughly 35 trillion words, seven times more than Russian and nine times more than Chinese, while Serbian, Latvian, Estonian and Slovenian fall below 30 billion and Irish and Maltese, both official EU languages, below two billion each, four orders of magnitude less than English.

Open-weight models are typically trained on such crawls, so their per-language competence largely inherits this distribution; matching English-level performance in the smallest languages evidently requires resources beyond these datasets. Correlating each model's clemscore with crawled volume confirms this: the association is strong for the open-weight models (Pearson $r = 0.78$ with $\log_{10}$ word count, Spearman $\rho = 0.72$, $p < 0.001$) and absent for the two commercial ones ($\rho = 0.03$ and $0.06$, both n.s.), which perform as well on the languages with the least crawled text as on those with the most.

\end{document}